\documentclass[lettersize,journal]{IEEEtran}

\usepackage{microtype}
\usepackage{graphicx}
\usepackage{soul}
\usepackage{xcolor}

\usepackage{textcomp}
\usepackage{stfloats}
\usepackage{verbatim}
\usepackage{cite}
\usepackage{caption}
\usepackage{subcaption}
\usepackage{algorithm2e}
\RestyleAlgo{ruled}
\usepackage{booktabs}
\usepackage{orcidlink}
\usepackage{hyperref}
\definecolor{mydarkgreen}{rgb}{0,0.6,0.30}
\definecolor{mydarkblue}{rgb}{0,0.30,0.65}
\hypersetup{
    colorlinks=true,    
    linkcolor=mydarkblue,     
    citecolor=mydarkgreen,    
    urlcolor=mydarkblue       
}

\usepackage{amsmath}
\usepackage{amssymb}
\usepackage{mathtools}
\usepackage{amsthm}
\usepackage{amsfonts}

\usepackage[capitalize,noabbrev]{cleveref}

\def\rvA{{\mathbf{A}}}

\def\rvF{{\mathbf{F}}}
\def\rvD{{\mathbf{D}}}
\def\rvH{{\mathbf{H}}}

\def\rvz{{\mathbf{z}}}
\def\rvq{{\mathbf{q}}}
\def\rvy{{\mathbf{y}}}
\def\rvx{{\mathbf{x}}}
\def\rvn{{\mathbf{n}}}
\def\rvu{{\mathbf{u}}}
\def\rvv{{\mathbf{v}}}
\def\rvf{{\mathbf{f}}}

\def\h{{\mathbf{h}}}

\def\p{{\mathbf{p}}}

\newcommand{\R}{\mathbb{R}}
\DeclareMathOperator*{\argmin}{argmin}

\begin{document}

\title{LoFi: Neural Local Fields for Scalable Image Reconstruction}

\author{AmirEhsan Khorashadizadeh$^{\orcidlink{0000-0003-0660-6823}}$,
Tobías I. Liaudat$^{\orcidlink{0000-0002-9104-314X}}$,
Tianlin Liu$^{\orcidlink{0000-0001-8231-7594}}$,
Jason D. McEwen$^{\orcidlink{0000-0002-5852-8890}}$,
and Ivan Dokmani\'c$^{\orcidlink{0000-0001-7132-5214}}$
\thanks{
AmirEhsan Khorashadizadeh, Tianlin Liu and Ivan Dokmani\'c were supported by the European Research Council Starting Grant 852821--SWING.

AmirEhsan Khorashadizadeh is also supported by the Promotion of Young Talents at the University of Basel.

AmirEhsan Khorashadizadeh is with the Department of Mathematics and Computer Science of the University of Basel, 4001 Basel, Switzerland and also with the Mullard Space Science Laboratory (MSSL), University College London (UCL), Surrey RH5 6NT, UK (e-mail: \href{mailto:amir.kh@unibas.ch}{amir.kh@unibas.ch}).

Tianlin Liu is with the Department of Mathematics and Computer Science of the University of Basel, 4001 Basel, Switzerland (e-mail: \href{mailto:t.liu@unibas.ch}{t.liu@unibas.ch}).

Tobías I. Liaudat is with the IRFU, CEA, Université Paris-Saclay, F-91191 Gif-sur-Yvette, France (e-mail: \href{mailto:tobias.liaudat@cea.fr}{tobias.liaudat@cea.fr}).

Jason D. McEwen is with the Mullard Space Science Laboratory (MSSL), University College London (UCL), Surrey RH5 6NT, UK (e-mail: \href{mailto:jason.mcewen@ucl.ac.uk}{jason.mcewen@ucl.ac.uk}).

Ivan Dokmani\'c is with the Department of Mathematics and Computer Science of the University of Basel, 4001 Basel, Switzerland, and also with the Department of Electrical, Computer Engineering, the University of Illinois at Urbana-Champaign, Urbana, IL 61801 USA (e-mail: \href{mailto: ivan.dokmanic@unibas.ch}{ivan.dokmanic@unibas.ch}).

Our implementation is available at \url{https://github.com/AmirEhsan95/LoFi}.

}
}

\maketitle
\begin{abstract}

Neural fields or implicit neural representations (INRs) have attracted significant attention in computer vision and imaging due to their efficient coordinate-based representation of images and 3D volumes. In this work, we introduce a coordinate-based framework for solving imaging inverse problems, termed LoFi (Local Field). Unlike conventional methods for image reconstruction, LoFi processes local information at each coordinate \textit{separately} by multi-layer perceptrons (MLPs), recovering the object at that specific coordinate. Similar to INRs, LoFi can recover images at any continuous coordinate, enabling image reconstruction at multiple resolutions. With comparable or better performance than standard deep learning models like convolutional neural networks (CNNs) and vision transformers (ViTs), LoFi achieves excellent generalization to out-of-distribution data with memory usage almost independent of image resolution. Remarkably, training on $1024 \times 1024$ images requires less than 200MB of memory---much below standard CNNs and ViTs. Additionally, LoFi's local design allows it to train on extremely small datasets with 10 samples or fewer, without overfitting and without the need for explicit regularization or early stopping.
\end{abstract}

\section{Introduction}
\label{sec: introduction}

\IEEEPARstart{I}{maging} inverse problems are ubiquitous in domains like medicine~\cite{wang2008outlook}, material science~\cite{holler2017high}, remote sensing~\cite{blahut2004theory} and cosmology~\cite{kaiser1993}. Deep learning is often the method of choice for solving inverse problems. In particular, convolutional neural networks (CNNs) like U-Net~\cite{ronneberger2015u} have shown remarkable performance across diverse tasks including computed tomography (CT)~\cite{jin2017deep}, magnetic resonance imaging (MRI)~\cite{hyun2018deep}, radio interferometric imaging \cite{mars2024learned} and dark matter mapping~\cite{jeffrey2020deep} in cosmology. The success of the U-Net is primarily attributed to its multiscale architecture with a large receptive field that extracts features from the input image at different scales~\cite{liu2022learning}.

Despite their strong performance on low-dimensional 2D imaging problems, deep learning architectures can become computationally expensive for large images~\cite{zhang2017beyond, liang2021swinir, wang2021pyramid, sepehri2024serpent}. This inefficiency arises because current deep neural networks reconstruct the entire image simultaneously, necessitating substantial memory for back-propagation during training, especially for large images. Additionally, interpreting such complex architectures~\cite{zhang2021plug,liang2021swinir, fabian2022humus} and analyzing their reconstructions can be challenging. Simplified neural architectures can enhance our understanding of reconstruction mechanisms, enabling the design of robust models with strong generalization~\cite{aggarwal2018modl}.

Recently, implicit neural representations (INRs)~\cite{sitzmann2020implicit, atzmon2020sal, chabra2020deep} have emerged as a promising coordinate-based pipeline for representing continuous signals, images, and 3D volumes. Unlike most existing deep learning models that treat signals as discrete arrays, INRs map coordinates to signal values using a deep neural network, typically a multi-layer perceptron (MLP), resulting in a \textit{continuous} signal representation. INRs have several advantages over standard deep learning models. Rather than representing signals at a single resolution, INRs can conveniently interpolate signals in a continuous space. This coordinate-based representation is particularly interesting because we can adjust the required memory making INRs well-suited for high-dimensional 3D reconstructions \cite{chen2019learning, peng2020convolutional, jiang2020local, dupont2022data, dupont2021generative, susmelj2024uncertainty} and scene representation \cite{mildenhall2021nerf}.

In this paper, we introduce a scalable coordinate-based local reconstruction pipeline for solving imaging inverse problems. In many tasks such as image denoising and low-dose computed tomography (LDCT), the image intensity at a specific pixel is primarily influenced by the observed image in the pixel’s neighborhood~\cite{mohan2019robust}. Our proposed model, termed LoFi (Local Field), recovers the image intensity at each pixel \textit{separately}, using local information extracted from the input image around that pixel. This information is processed by a neural network (composed of MLP blocks) to determine the image intensity at the target pixel, enabling image reconstruction at any resolution or arbitrary continuous coordinate. 

\begin{figure*}
\centering
\begin{subfigure}{0.5\textwidth}
  \centering
 \includegraphics[width=1.05\textwidth]{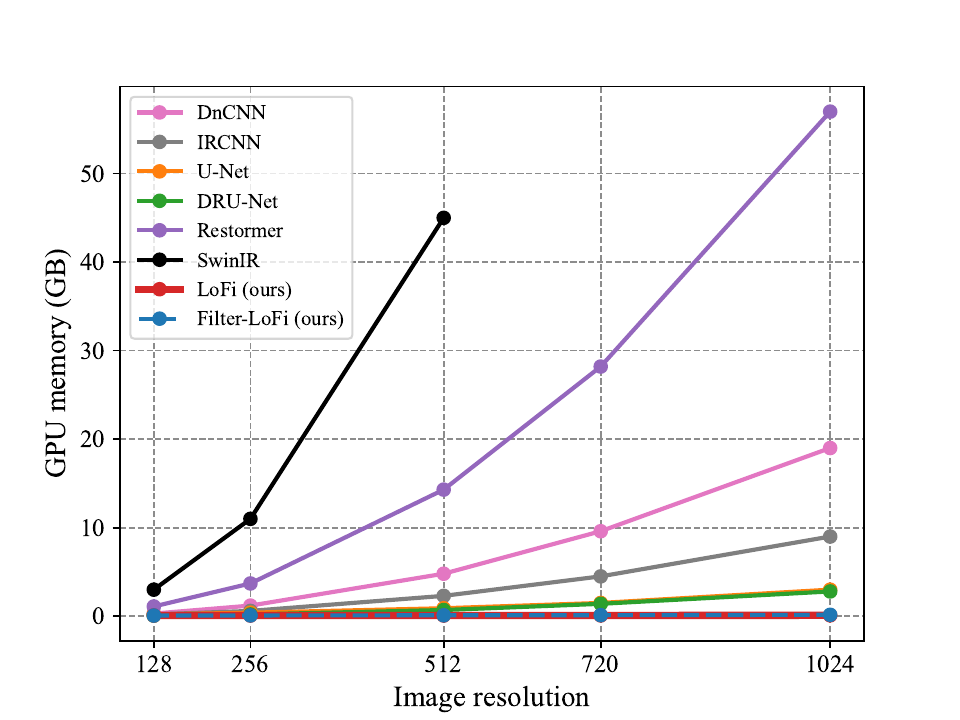}
\caption{Memory footprint  (batch size 1)}
\label{fig: gpu_train}
\end{subfigure}%
\begin{subfigure}{0.5\textwidth}
\centering
\includegraphics[width=1.05\textwidth]{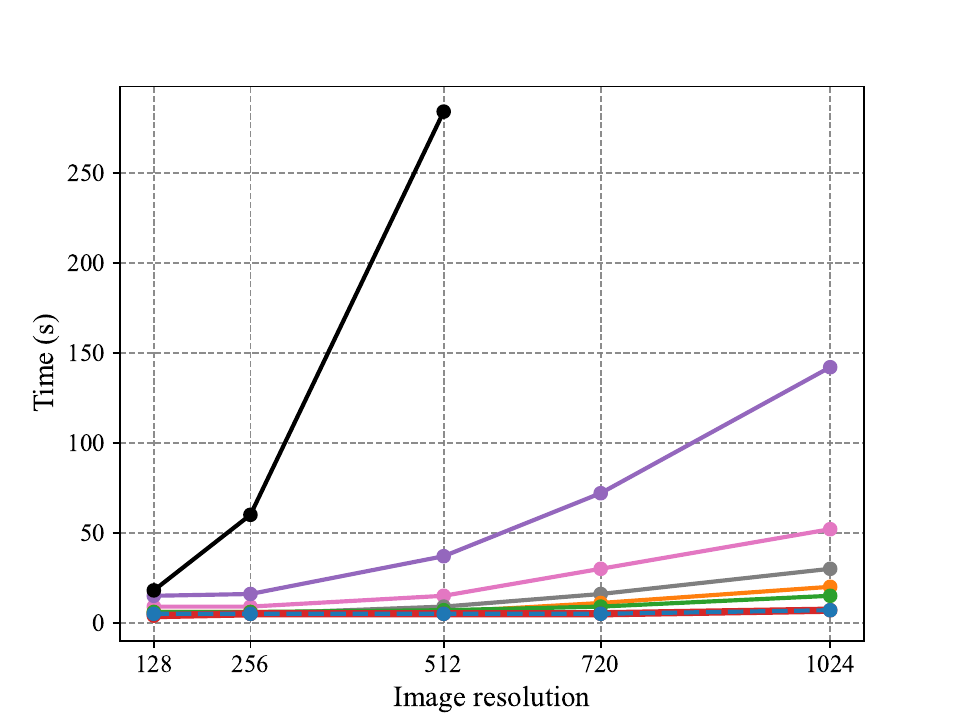}
\caption{Training time (100 iterations)}
\label{fig: time_train}
\end{subfigure}
\caption{The memory and time requirements during training for different models; LoFi is significantly faster and more memory-efficient than CNNs and ViTs. Notably, LoFi's memory usage remains almost independent of image resolution, making it an ideal choice for high-dimensional image reconstruction. All experiments are conducted using a single A100 GPU with 80GB memory. Missing data points indicate that the corresponding model exceeds the GPU memory capacity for the given resolution.}
\label{fig: training computation}
\end{figure*}

LoFi's coordinate-based design brings several advantages. Unlike standard CNNs and ViTs which requires substantial memory during training, we can train LoFi, Similar to INRs, on mini-batches of both objects and pixels. This results in resolution-agnostic memory usage, making it highly efficient for high-resolution image processing. As illustrated in Figure \ref{fig: training computation}, LoFi demonstrates significantly lower memory and time requirements compared to CNNs and ViTs.
For instance, training LoFi on $1024 \times 1024$ images requires less than 200MB memory. Furthermore, its local design provides a strong inductive bias for image reconstruction; unlike CNNs and ViTs, which tend to overfit when trained on tiny datasets with very few samples, LoFi consistently shows robust performance without overfitting, eliminating the need for regularization or early stopping.

In addition, deep learning models for image reconstruction like multiscale CNNs~\cite{ronneberger2015u, zhang2021plug} and vision transformers (ViTs)~\cite{liang2021swinir, wang2022uformer, fabian2022humus, zhao2023comprehensive} often use sophisticated architectures that complicate downstream interpretations.
Through our proposed coordinate-conditioned patch geometry, LoFi can learn the position of the relevant features in the input image for each pixel, providing insights for downstream image analysis and interpretation.

\begin{figure*}[t]
    \centering
    \includegraphics[width = \textwidth]{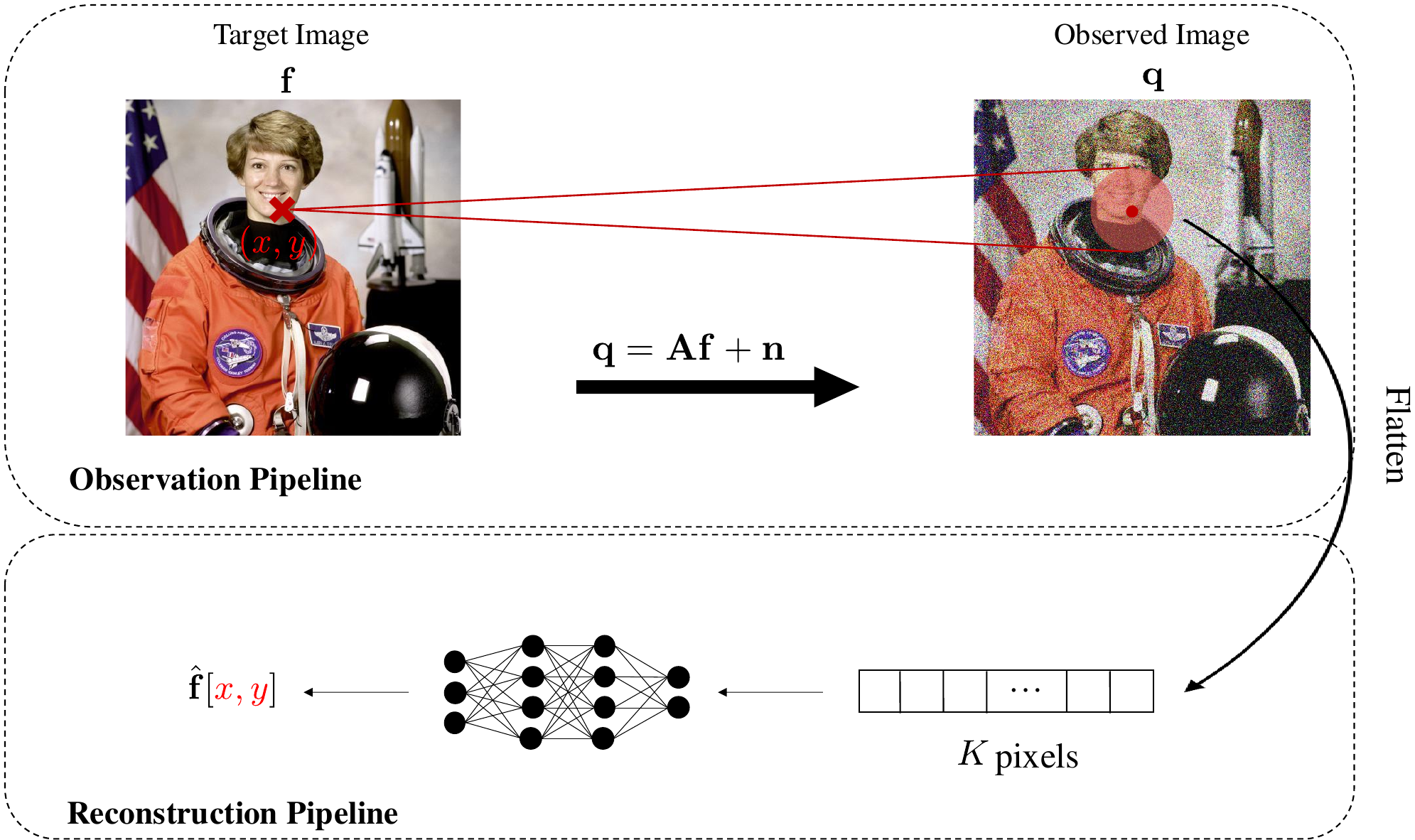}
    \caption{LoFi; the neural network $\text{NN}_\theta$, typically composed of MLP modules, processes the local information extracted from the observed image around the given pixel $(x,y)$. LoFi's inductive bias for image reconstruction brings strong generalization on OOD data, it requires less training data and uses small memory when training on high-resolution images.}
    \label{fig: model}
\end{figure*}

\section{Background and related works}
\label{sec: background}
The goal of an imaging inverse problem is to recover the image of interest $\rvf$ from noisy measurements $\rvq$ acquired via the potentially non-linear forward operator $\rvA$,
\begin{equation}
    \rvq = \rvA(\rvf) + \rvn,
    \label{eq: forward model}
\end{equation}
where $\rvn$ is noise. Inverse problems are often ill-posed, implying the existence of multiple plausible reconstructions for a given set of measurements. It is thus important to incorporate an image prior to enable stable and accurate reconstruction.
While traditional hand-crafted priors and regularizers such as total variation (TV) \cite{osher2005iterative} can yield good reconstructions, learning-based priors have brought about substantial improvements~\cite{kothari2021trumpets, jalal2021robust, chung2022diffusion, hu2023restoration}.

\subsection{Inverse problems with local operators}
\label{sec: local_ops}
In many inverse problems, the forward operator is local; an image pixel primarily relates to its neighboring pixels in the observed image. Here the assumption is that the observed data is shaped like an image with the same coordinate system as the target. A local forward operator suggests an opportunity to design efficient deep-learning architectures. In the following, we discuss three inverse problems characterized by this localization property.

\paragraph{Image denoising}

The most straightforward inverse problem is image denoising, where the forward operator $\rvA$ in \eqref{eq: forward model} is the identity matrix; each pixel in the clean image is solely connected to the corresponding pixel in the noisy image, assuming independent noise (between pixels) and natural images \cite{mohan2019robust}. Despite this simplicity, the presence of noise prompts the incorporation of `contextual' information from neighboring noisy pixels. In principle, the required local neighborhood size depends on the noise level and the prior distribution; stronger noise usually calls for larger receptive fields~\cite{levin2011natural, mohan2019robust}.

\paragraph{Low-dose computed tomography}

Low-dose computed tomography (LDCT) is a medical imaging technique that reduces radiation exposure to patients during a CT scan by lowering the X-ray beam intensity \cite{mayo1997simulated}. Traditional CT imaging uses X-rays to capture cross-sectional images of the body, but these scans can result in significant radiation doses, potentially increasing the risk of long-term side effects like cancer~\cite{brenner2007computed, einstein2007estimating}. LDCT mitigates these risks by significantly lowering the X-ray dose without compromising diagnostic quality, making it a safer imaging process.

CT image reconstruction can be viewed as an inverse problem,
\begin{equation}
    \rvq = \mathcal{R} (\rvf) + \rvn,
\end{equation}
where the forward operator $\mathcal{R}$ is the radon transform \cite{helgason1980radon} applied in angles $\{\alpha_i\}_{i=1}^r$ and $\rvn$ is the additive noise. In LDCT, the projections $\rvy$ are collected from sufficient angles $r$ but using low-intensity beams resulting in noisy projections. The standard approach for CT image reconstruction is filtered backprojection (FBP),
\begin{equation}
    \rvf_\text{FBP} = \mathcal{R}_\text{FBP}(\rvq).
\end{equation}
In an ideal scenario where we have an infinite number of noise-free projections, FBP is the exact inverse, namely, $\rvf_\text{FBP} = \rvf$. In LDCT, we have a sufficient number of noisy projections $N$, the image reconstruction from FBP can be seen as a denoising problem,
\begin{equation}
    \rvf_\text{FBP} \approx \rvf + \mathcal{R}_\text{FBP}(\rvn).
    \label{eq: FBP}
\end{equation}
The image reconstruction from FBP can be seen as an inverse problem with a local forward operator for LDCT. For a comprehensive evaluation of current LDCT image reconstruction methods, we refer to \cite{kiss2024learned}, where the authors conducted an in-depth analysis of the performance of CNNs, using both simulated and real experimental data for training and inference.

\paragraph{Dark matter mapping in cosmology}

As another example, we consider mapping the dark matter distribution of the Universe \cite{kaiser1993} by solving a weak gravitational lensing \cite{mandelbaum2018} inverse problem. We will study the convergence that traces the projected mass distribution of dark matter in the Universe. The goal is to estimate the target convergence field $\kappa$ that traces the projected mass distribution, from a related observable, the shear field $\gamma$. Higher order statistical properties of the convergence field are highly informative for studying the nature of dark energy and dark matter. The two fields are related as,
\begin{equation}
    \gamma = \rvA \kappa + \rvn,
    \label{eq:mass_map_forward_op}
\end{equation}
where $\rvn$ is the shear field noise, and the forward operator $\rvA$ is a convolutional filter (a Fourier multiplier) with the 2D Fourier transform given by,\footnote{This is known as the weak lensing planar forward model. Note that $\rvD_{k_x, k_y}$ is undefined for $(k_x,k_y)=(0, 0)$, meaning that the observable $\gamma$ cannot constrain the average of $\kappa$. This is known as the mass sheet degeneracy. The choice $\rvD_{0, 0,} = 1$ is made for convenience and is secondary to what follows.}
\begin{equation}
\rvD_{k_x, k_y} = 
        \frac{k_{x}^{2} - k_{y}^{2} + 2 i k_x k_y}{k_{x}^{2} + k_{y}^{2}}, \quad \text{for } k_x^2 + k_y^2 \neq 0.
\end{equation}
A simple approach to recover the convergence field $\kappa$ from the shear field $\gamma$ is the Kaiser-Squires (KS) method~\cite{kaiser1993}, the current standard in cosmology. The method consists of directly applying $\rvA^{-1}$, the inverse of the forward operator, to the shear field, followed by a subsequent Gaussian smoothing with an ad hoc smoothing scale. We consider the naive KS inversion, without the Gaussian smoothing step, as follows,
\begin{equation}
    \kappa_\text{KS} = \rvA^{-1}\rvA \gamma =  \kappa + \rvA^{-1}\rvA \rvn = \kappa + \tilde{\rvn},
    \label{eq: KS}
\end{equation}
where $\tilde{\rvn} = \rvA^{-1}\rvA \rvn$. The recovery of the convergence field from the naive inverted KS image is thus an inverse problem with a local forward operator. We note that the processed noise $\tilde{\rvn}$ is now spatially correlated and modeling it may still benefit from a larger receptive field; this is addressed by the learnable noise suppression filter described in Section \ref{sec: filter}.

\subsection{Deep learning for image reconstruction}
Over the last decade, various deep learning approaches have achieved state-of-the-art (SOTA) performance in image reconstruction. Some early approaches segmented the image into patches and employed an MLP to denoise each patch individually~\cite{burger2012image}. However, the optimal patch size for high-quality reconstructions depends on the noise level and data distribution \cite{mohan2019robust}. Zhang et al.~\cite{zhang2017beyond} introduced DnCNN, an image-to-image CNN designed for image denoising where the network depth is adjusted based on the required receptive field size suggested by traditional methods~\cite{zoran2011learning}. To expand CNN's receptive field and use more contextual information, they used dilated convolutions \cite{zhang2017learning}. Concurrently, the U-Net~\cite{ronneberger2015u} has been the backbone model for SOTA image reconstruction architectures~\cite{liu2018multi, park2019densely, jia2021ddunet, liu2022learning}, in particular, the DRU-Net~\cite{zhang2021plug} achieved remarkable performance by integrating residual blocks~\cite{he2016deep} into the U-Net architecture.

More recently, vision transformers (ViTs)~\cite{dosovitskiy2020image} have achieved comparable or even better performance than CNNs for various image reconstruction tasks~\cite{chen2021pre, liang2021swinir, wang2022uformer, mansour2022image}. The core concept of ViTs is to divide the input image into non-overlapping patches, transform each patch into an embedding using a learned linear projection, and then process the resulting tokens through a series of stacked self-attention layers and MLP blocks. The advantage of ViTs over CNNs is attributed to their ability to capture long-range dependencies in the images through the self-attention mechanism \cite{liang2021swinir, zhao2023comprehensive}. As shown in Figure \ref{fig: training computation}, a major challenge for ViTs in image reconstruction is the quadratic computational complexity relative to the input image size~\cite{wang2021pyramid, zamir2022restormer} which makes ViTs impractical for high dimensional image reconstruction \cite{sepehri2024serpent}. Recent works attempted to reduce the computational complexity of vision transformers. Restormer \cite{zamir2022restormer} applies the self-attention along the channel dimension to reduce memory footprint. Some approaches design hierarchical transformers \cite{heo2021rethinking, wang2021pyramid}. The authors of \cite{fabian2022humus} use convolutional blocks to reduce the spatial dimensions followed by self-attention blocks applied in different scales. Despite the marginally better reconstructions compared to CNNs for inverse problems with non-local forward operators like magnetic resonance imaging (MRI), they show comparable performance with CNNs for local problems like image denoising \cite{fabian2022humus}.

In this paper, we demonstrate that for inverse problems with local forward operators, such as those discussed in Section~\ref{sec: local_ops}, it is not necessary to process large-scale features using multiscale CNNs or ViTs. Instead, our proposed architecture efficiently captures local features, resulting in strong generalization while significantly reducing computational costs and memory requirements, well-suited for high-dimensional image reconstruction.

\subsection{Neural fields for inverse problems}
INRs, particularly those with sinusoidal activations, efficiently model signals and their spatial derivatives. This is particularly useful for solving partial differential equations (PDEs) \cite{sitzmann2020implicit, vlavsic2022implicit}. In the domain of continuous super-resolution, INRs have been effectively combined with CNNs. For instance, the authors of \cite{chen2021learning} proposed conditioning the INR's generation process on local features extracted by a CNN from the low-resolution image. Closer to our work, the authors of \cite{khorashadizadeh2022funknn} used a similar approach, where local features extracted from a low-resolution image are processed with a lightweight CNN to enable continuous super-resolution.

\subsection{Patch-based image reconstruction}
While traditional methods for image reconstruction often process small patches extracted from the input image \cite{buades2005non, dabov2007image}, deep learning models with patch-based design have recently gained considerable attention. ViTs~\cite{dosovitskiy2020image} and MLP-mixers~\cite{tolstikhin2021mlp} are a well-known family of patch-based models whose strong performance is often attributed to patch-based design~\cite{trockman2022patches}. Deep generative models trained on small patches have been used for high dimensional image generation \cite{ding2023patched} and more efficient training on datasets with significantly fewer images \cite{wang2024patch}. Patch priors \cite{gilton2019learned, altekruger2022patchnr, hertrich2022wasserstein, piening2024learning} as efficient alternatives to image priors have been used for image reconstruction which allows training on small datasets with few images well-suited for scientific applications where we have access to a limited number of ground truth images.

\section{LoFi: a local field processor}
\label{sec: LoFi}
Our objective is to reconstruct the target image $\rvf \in \mathbb{R}^{N \times N \times C}$ from the observed image $\mathbf{q} \in \mathbb{R}^{M \times M \times C}$, as described in the forward equation \eqref{eq: forward model}, where $C$ is the channel dimension, $N$ and $M$ denote the resolution of the target and observed images. The resolutions of $\rvf$ and $\rvq$ may be different but we assume that they share the same coordinate system and semantics; thus $\rvf[x, y]$ and $\rvq[x, y]$ refer to the same continuous location in both images. Example problems where this is meaningful are denoising and deblurring. More generally, for many linear inverse problems, including computed tomography, it will hold after applying the adjoint of the forward operator as shown in \eqref{eq: FBP}.

We then consider a local forward operator $\rvA$ where the intensity $\rvf[x,y]$ at pixel $(x,y) \in \R^2$ relates only to a small neighborhood of $(x, y)$ in $\rvq$. Based on this locality hypothesis, we design a coordinate-based neural network inspired by implicit neural representations. 

To recover the image (only) at the target pixel $(x,y)$, we extract a neighborhood of $K$ pixels around $(x,y)$ from the observed image $\rvq$. We denote the (flattened) extracted pixels by $\p_{x,y} \in \R^{K \times C}$. As illustrated in Figure~\ref{fig: model},
we process $\p_{x,y}$ using a neural network $\text{NN}_\theta: \R^{K \times C} \to \R^C$ with parameters $\theta$ to approximate the image intensity $\rvf[x,y]$,
\begin{equation}
    \hat{\rvf}[x,y] = \text{NN}_\theta (\p_{x,y}).
    \label{eq: LoFi reconstrction}
\end{equation}
This coordinate-based image representation enables image recovery at any continuous coordinate with small memory. We call our framework LoFi (Local Field). As we typically use a small neighborhood size $K$, we can parametrize $\text{NN}_\theta$ using an MLP. In the following sections, we provide further details regarding the LoFi architecture and introduce the pre-processing filter, patch geometry, and training strategy.

\subsection{MultiMLPs for large neighborhoods}
\label{sec: multi-MLP}
\begin{figure}
    \centering
    \includegraphics[width = 0.43 \textwidth]{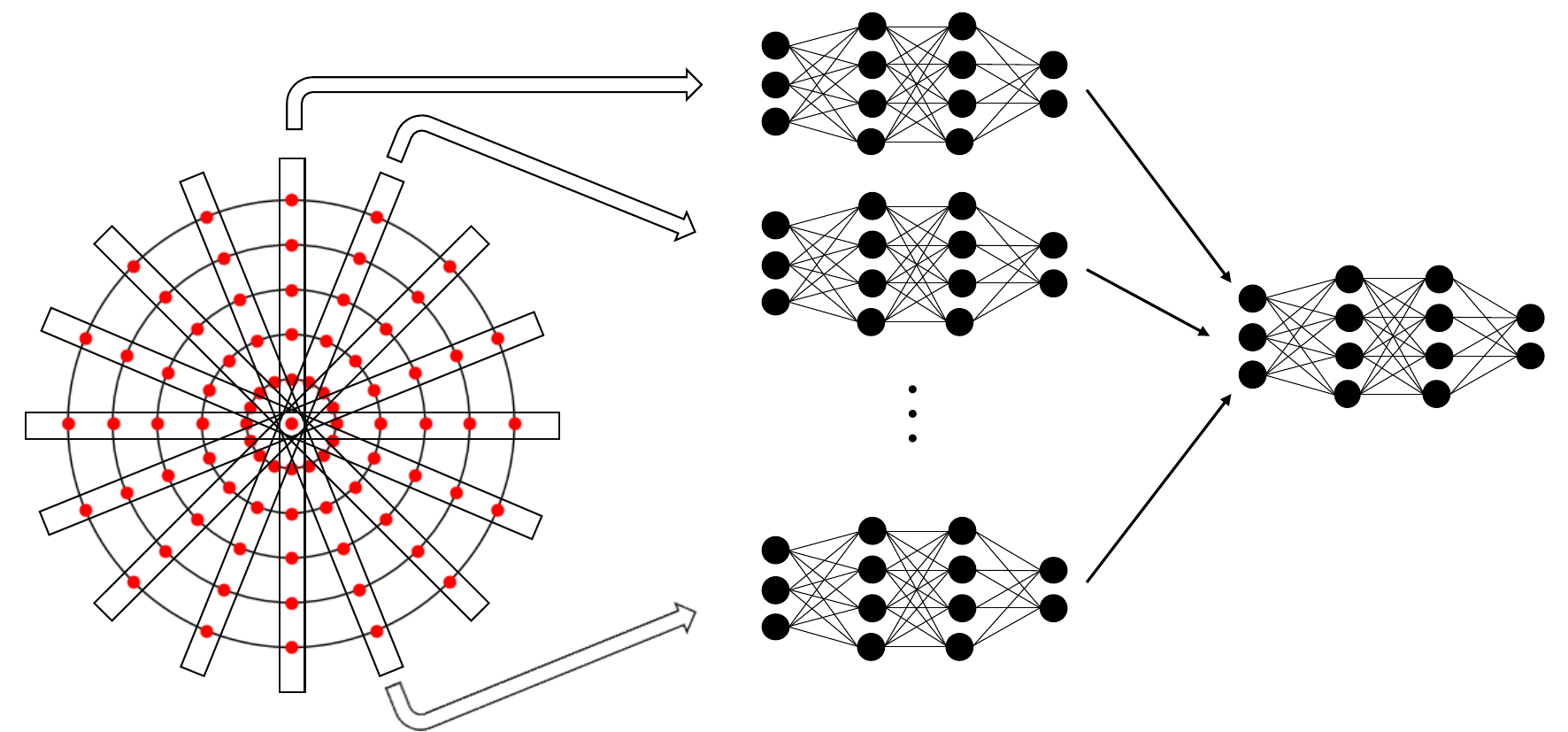}
    \caption{MultiMLP architecture; the input information is split into smaller chunks each processed with a separate MLP, the extracted information is then mixed by another MLP.}
    \label{fig: multiMLP}
\end{figure}
One important factor in LoFi architecture is the neighborhood dimension $K$; larger $K$ can provide more detailed information to improve reconstructions. However, the size of the MLP quadratically scales with $K$ resulting in large networks and slow training. To address this issue, we propose MultiMLP, an alternative architecture to handle large neighborhoods. Inspired by vision transformers~\cite{dosovitskiy2020image}, we split the extracted neighboring pixels (here over a circular geometry) into smaller chunks, each processed by a separate MLP as shown in Figure~\ref{fig: multiMLP}. Their outputs are then mixed by another MLP.

\subsection{Patch geometry and equivariance}
\label{sec: patch geo}

A straightforward method to extract the neighboring pixels is to directly select the pixels adjacent to the target pixel $(x, y)$. This however has several drawbacks: it requires a manual selection of the neighborhood size, akin to traditional methods, which is hard to estimate; it only permits image evaluation at on-grid pixels resulting in a single-resolution image reconstruction.

\subsubsection{Differentiable patch extraction}
While we could use the standard square patch to extract the local information, this would not result in a differentiable map from the input coordinate to the pixel intensity. To enable image reconstruction at arbitrary \textit{continuous} coordinate, we extract $\p_{x,y}$ using bicubic interpolation which results in a differentiable architecture.

\subsubsection{Learnable patch geometry}

This differentiable patch extraction opens up the possibility to learn the patch geometry centered around the target coordinate $(x,y)$ by first defining a set of \textit{learnable} coordinate offsets,
\begin{equation}
    \Delta I = \big[(x^\Delta_n, y^\Delta_{n^\prime})\big]_{n,n^\prime = 1}^K.
    \label{eq: learnable patch}
\end{equation}
By learning the optimal coordinate offsets $\Delta I$ we can also control the scale and shape of the patch receptive field. We then extract the patch by evaluating $\rvq$ at coordinates $I_{(x, y)} = (x, y) + \Delta I$ to get
\begin{equation}
    \p_{x,y} = \rvq[I_{(x,y)}].
\end{equation}
We emphasize that the coordinates $I_{(x, y)}$ need not be aligned with the pixel grid as we use differentiable bicubic interpolation to enable off-the-grid evaluation. Unlike most other work using square patches, we initialize $\Delta I$  to a circular geometry; this results in better rotation invariance as discussed in Section \ref{sec: rot_equi} in the supplementary materials.

\subsection{Coordinate-conditioned patch geometry}

\label{sec: Coordinate-based patch geometry}
The learned patch geometry in \eqref{eq: learnable patch} is fixed. In practice, the relative locations of the most relevant information changes for different pixels. To enable coordinate-conditioned patch geometry (CCPG), we use another neural network (typically a simple MLP) $\text{CCPG}_\psi: \R^{K \times C} \to \R^{2K}$ which takes a local patch around pixel $(x,y)$ and estimates the position of the coordinates inside the patch as follows,
\begin{equation}
    \Delta I_{CCPG} (x,y) = \text{CCPG}_\psi (p_{x,y}).
\end{equation}
We then evaluate $\rvq$ at $(x,y) + \Delta I_{CCPG} (x,y)$. The learned patch geometry $\Delta I_{CCPG} (x,y)$ can localize the position of the relevant information for image recovery at pixel $(x,y)$ for corrupted image $\rvq$. A similar strategy has been developed by \cite{dai2017deformable} for object detection using CNNs.

Now we can improve the localization performance by repeating this process $T$ times,
\begin{align}
    &\Delta I_{CCPG}^{(i)} (x,y) = \text{CCPG}_\psi^{(i)} (p_{x,y}^{(i-1)}) \\
    &p_{x,y}^{(i)} = \rvq[(x,y) + \Delta I_{CCPG}^{(i)} (x,y) ]
\end{align}
where in each iteration $1 \leq i \leq T$, we feed noisy image $\rvq$ evaluated at the estimated patch position in the previous step $p_{x,y}^{(i-1)}$ to a separate neural network $\text{CCPG}_\psi^{(i)}$. As we will show in our experiments in Section~\ref{sec: adaptive patch experiment}, the learned patch geometry indeed improves over iterations.

\subsection{Noise suppression filter}
\label{sec: filter}

The observed image $\rvq$ is corrupted with noise. Optimal filters for correlated noise can have a large receptive field which cannot be implemented by the localized $\text{NN}_\theta$ in \eqref{eq: LoFi reconstrction}. To address this, we apply pre-processing convolutional filters $\h \in \R^{s \times s \times L}$ to the observed image before patch extraction,
\begin{equation}
    \tilde{\rvq} = \rvq \star \h,
    \label{eq: conv filter}
\end{equation}
where $\star$ denotes the convolution operator, $s$ is the filter size and $L$ is the number of filters. This process is reminiscent of filtering techniques commonly used in computed tomography imaging~\cite{diwakar2018}. We can also concatenate the filtered and noisy images along the channel dimension. 

To avoid manual adjustment of the receptive field, we apply the filter in the frequency domain,
\begin{equation}
    \tilde{\rvq} =  \rvF^{-1} \rvH \rvF \rvq,
    \label{eq: fourier filter}
\end{equation}
where $\mathbf{F}$ and $\mathbf{F}^{-1}$ are the forward and inverse 2D Fourier transforms, and the diagonal matrix $\rvH$ is a learnable parameter fitted to the data. The resulting filter can have large support in image space, capturing global information. We initialize the filter $\rvH$ with all ones in the Fourier domain (an identity filter) which corresponds to the smallest support in pixel space.
During training, the filter expands its support based on the image resolution, noise level, and data distribution.

\subsection{Resolution-agnostic memory usage in training}

\label{sec: training}
For simplicity, we write the entire LoFi pipeline as $\hat{\rvf}(\rvx) = \text{LoFi}_\phi (\rvx, \rvq)$ where $\rvx=(x,y)$ is the target pixel. The model approximates the image intensity $\hat{\rvf}(\rvx)$ from the observed image $\rvq$. We denote all the trainable parameters of LoFi by $\phi$. This includes $\theta$, pre-processing filter parameters $\h$, and CCPG weights $\psi$. We consider a set of training data $\{(\rvq_i, \rvf_i)\}_{i = 1}^D$ from observed and target images. We optimize the LoFi parameters $\phi$ using a gradient-based optimizer to solve,
\begin{equation}
    \phi^\star = \argmin_\phi \sum_{i=1}^D \sum_{j=1}^{N^2} | \rvf_i(\rvx_j) - \text{LoFi}_\phi (\rvx_j, \rvq_i) |.
    \label{eq: training loss}
\end{equation}
At inference time, we can recover the image at any target pixel $\rvx$ from the observed image $\rvq_\text{test}$ as $\hat{\rvf}_\text{test}(\rvx)=\text{LoFi}_{\phi^\star}(\rvx, \rvq_\text{test})$. From \eqref{eq: training loss}, it is clear that LoFi can be trained on mini-batches of both objects and pixels, enabling an almost constant-in-resolution memory usage during training and inference. This flexibility also allows us to train LoFi on datasets with images in various resolutions.

\section{Further Extensions and Applications}

\begin{figure*}[t]
    \centering\includegraphics[width = \textwidth]{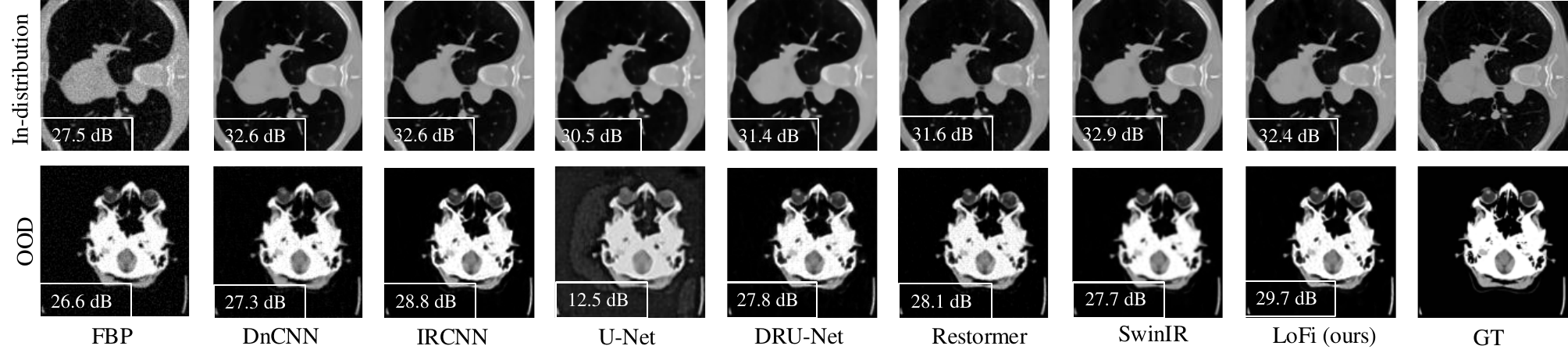}
    \\ [-5pt]
    \caption{Performance comparison on LDCT (30dB noise) at resolution $128 \times 128$ for in-distribution (chest) and OOD (brain) data.}
    \label{fig: results_CT}
\end{figure*}
In this section, we discuss further applications of the proposed local processing framework.

\subsection{Learning the imaging geometry}

Many seemingly complex image-to-image transforms, like those involving translations and rotations, alter image content using simpler coordinate transformations. For example, transposing a 2D image causes significant pixel differences in the high-dimensional pixel space; but the transformation relies on a simple, low-dimensional coordinate swap. When CNNs are fitted on image-to-image tasks, they are directly trained to model the changes in pixel values and, therefore, tend to overlook the coordinate transformations that underlie the pixel value changes. This is particularly problematic for large-scale changes in the coordinate system as the CNN should have a large receptive field to capture these geometric changes. This limitation has been explored by existing works to build deep neural networks that can model the changes in the coordinate system as well as the image content. For example, the authors of \cite{kothari2020learning} separately model changes of the coordinate system and the image content using CNNs.

Although LoFi can only learn the local changes in the \textit{image content}, its coordinate-based design allows us to easily equip it to learn the changes in the coordinate system as well. To this end, we use an $\text{INR}_\eta$ with parameters $\eta$, that takes the given coordinate $\rvx$ and returns the new coordinate $\tilde{\rvx}$. This INR is then followed by LoFi to sample from the local neighborhood around $\tilde{\rvx}$ from the input image $\rvq$ as follows,
\begin{align}
    \tilde{\rvx} &= \text{INR}_\eta (\rvx), \\
    \hat{\rvf}(\rvx) &= \text{LoFi}_\phi(\tilde{\rvx}, \rvq).
\end{align}
We train the above architecture end-to-end where the task of INR is to learn the changes in the coordinate system, while LoFi learns the changes in the image content from the corrected coordinate system. We call this new architecture INR-LoFi and compare its performance with CNNs for a toy experiment in Section \ref{sec: exp_INR_LoFi}.

\subsection{Low-resolution priors for high-resolution unsupervised image reconstruction}
\label{sec: LoFi-admm}
As shown in Section \ref{sec: training}, the proposed LoFi is a supervised learning framework and requires paired training data of measurements and corresponding target images. This is hard to obtain in many real scientific problems~\cite{jin2017deep, mccann2017convolutional, wei2018deep, khorashadizadeh2023conditional}. Additionally, supervised models require re-training even for small changes in the forward model. 

To address this issue, unsupervised methods learn a data-driven prior from only target images. Once trained, they can be used as a prior for solving any inverse problem using an iterative process. Unsupervised methods based on deep generative models \cite{bora2017compressed, kothari2021trumpets, kawar2022denoising, khorashadizadeh2023deep, liu2023optimization} and plug-and-play (PnP) \cite{venkatakrishnan2013plug, chan2016plug, romano2017little} show strong performance on various image reconstruction tasks.
We provide a brief overview of the PnP-ADMM framework in Section~\ref{sec: PnP ADMM} of the supplementary material.

In this section, we use a pre-trained LoFi denoiser as a prior in the plug-and-play (PnP) framework. This integration allows us to use a LoFi denoiser trained on low-resolution images to solve inverse problems at any higher resolution. Since we use the PnP-ADMM backbone we call the resulting scheme  LoFi-ADMM  (cf. Algorithm~\ref{alg: LoFi-ADMM}). Even though the LoFi denoiser is trained on low-resolution images, LoFi-ADMM can be applied at arbitrary higher resolutions thanks to LoFi's continuous image representation.
\begin{algorithm}[h!]
\SetAlgoLined
\KwIn{$\rvq$, $\rvA$, $\phi^*$}
\textbf{Parameter:} $\alpha$
\BlankLine
$\rvf = 0$, $\rvu = 0$, $\rvv = 0$;
 
\For{$k = 0$ \KwTo $K-1$}{
    $\rvf_k \leftarrow h(\rvv_{k-1}  - \rvu_{k-1}; \alpha)$;
    
    $\rvv_k(\rvx) \leftarrow \text{LoFi}_{\phi^*}(\rvx,\rvf_k - \rvu_{k-1}), \quad \forall \rvx \in \rvv_k$;
    
    $\rvu_k \leftarrow \rvu_{k-1} + (\rvf_k - \rvv_k)$;
    }
 \caption{LoFi-ADMM}
 \label{alg: LoFi-ADMM}
\end{algorithm}

\section{Experiments}
\label{sec: experiments}

We now present a suite of experimental results which illustrate the various aspects of LoFi. In Section \ref{sec: experiment_image_rec}, we compare LoFi with successful CNNs and ViTs on several image reconstruction tasks. We compare LoFi with CNNs and ViTs in terms of memory usage and training time in Section \ref{sec: computation}.
In Section \ref{sec: adaptive patch experiment}, we use our proposed method in Section~\ref{sec: Coordinate-based patch geometry} for analysis and interpretation of the reconstructed image. In Section \ref{sec: exp_INR_LoFi}, we showcase the success of INR-LoFi and the failure of CNNs on a simple toy experiment where we have changes in the coordinate system. In Section \ref{sec: pnp_experiment}, we apply our unsupervised framework LoFi-ADMM to two inverse problems, image in-painting and radio interferometry.

\subsection{Scalable image reconstruction with LoFi}
\label{sec: experiment_image_rec}

We assess the performance of LoFi on inverse problems with local operators including low-dose computed tomography (LDCT) upon applying FBP, image denoising, and dark matter mapping in cosmology. As baselines, we consider two standard single-scale CNNs, DnCNN~\cite{zhang2017beyond} and IRCNN~\cite{zhang2017learning} and two multiscale CNNs, U-Net~\cite{ronneberger2015u} and DRU-Net~\cite{zhang2021plug}. We also consider two strong vision transformers, Restormer \cite{zamir2022restormer} and SwinIR \cite{liang2021swinir}, which achieved state-of-the-art performance on various image reconstruction problems. We assess the reconstruction quality with peak signal-to-noise ratio (PSNR) and structural similarity index (SSIM)~\cite{wang2004image}. Further details regarding the network architectures and training details are given in Section~\ref{sec: network architecture} in the supplementary materials.

\paragraph{Low-dose computed tomography (LDCT)}

We simulate parallel-beam LDCT with $r = 180$ uniformly distributed projection orientations. We add white Gaussian noise to obtain the signal-to-noise (SNR) ratio of 30dB.
We use 1000 training samples of chest images from the LoDoPaB-CT dataset~\cite{leuschner2021lodopab} at resolution $128 \times 128$. All models take FBP as input and return improved reconstruction in the output. The model performance is assessed on 64 in-distribution test samples of chest images. We additionally use 16 brain CT samples \cite{hssayeni2020computed} to evaluate model generalization on out-of-distribution (OOD) data.

The upper row of Figure~\ref{fig: results_CT} illustrates the performance of different models on in-distribution chest images. We see that LoFi achieves performance comparable to strong CNNs and ViTs while using only simple MLP modules. The bottom row shows model performance on OOD brain samples. It is remarkable that LoFi significantly outperforms CNNs and ViTs, indicating the strong generalization achieved by the locality of the architecture. Quantitative results are presented in Table \ref{tab: quantitative results}.

\begin{table}
\renewcommand{\arraystretch}{1.3}
      \centering
    \caption{Model comparison on the reconstruction quality for LDCT (30dB SNR) averaged on 64 in-distribution chest and 16 OOD brain samples.}
    \label{tab: quantitative results}
    \resizebox{0.5\textwidth}{!}{%
    \begin{tabular}{l |cc| cc}
    \hline
    & \multicolumn{2}{c|}
    {\textbf{In-dist (chest)}} & \multicolumn{2}{c}{\textbf{OOD (brain)}} \\
 \hline
  & \textbf{PSNR} & \textbf{SSIM} &  \textbf{PSNR} & \textbf{SSIM} \\
    \hline
    FBP   & 26.0 & 0.55 & 26.0 & 0.61\\
    \hline
    DnCNN~\cite{zhang2017beyond} & 34.2  & 0.90 &  27.6 & 0.82  \\
    IRCNN~\cite{zhang2017learning} & \textbf{34.5}  & \textbf{0.91} & 29.8  & 0.93  \\
    \hline
    U-Net~\cite{ronneberger2015u}   & 33.0  & 0.90 & 13.9 & 0.81 \\
    DRU-Net~\cite{zhang2021plug} &  33.8 & 0.90 &  29.0 & 0.90 \\
    \hline
    Restormer ~\cite{zamir2022restormer} &  33.5 & 0.89 &  28.9 & 0.79 \\
    SwinIR ~\cite{liang2021swinir} &  \textbf{34.5} & 0.90 &  29.3 & \textbf{0.94} \\
    \hline
    LoFi (ours) & 34.1 &  0.90  & \textbf{30.6} & \textbf{0.94}  \\
    \hline
    \end{tabular}
    }
\end{table}

\paragraph{Image denoising with tiny datasets}

Here we train LoFi on image denoising on a small dataset of only 10 images at resolution $512 \times 512$; cf. Figure~\ref{fig: training_set} in the supplementary materials. 9 images are used for training and 1 for evaluation. The images are scaled between 0 and 1 and we add white Gaussian noise with 
$\sigma=0.15$. We compare LoFi with CNN and 
Restormer models with a comparable number of trainable parameters (3M). SwinIR was excluded from this experiment due to its high computational cost and very long processing time at this resolution.

\begin{figure}
	\centering
    \includegraphics[width=\linewidth]{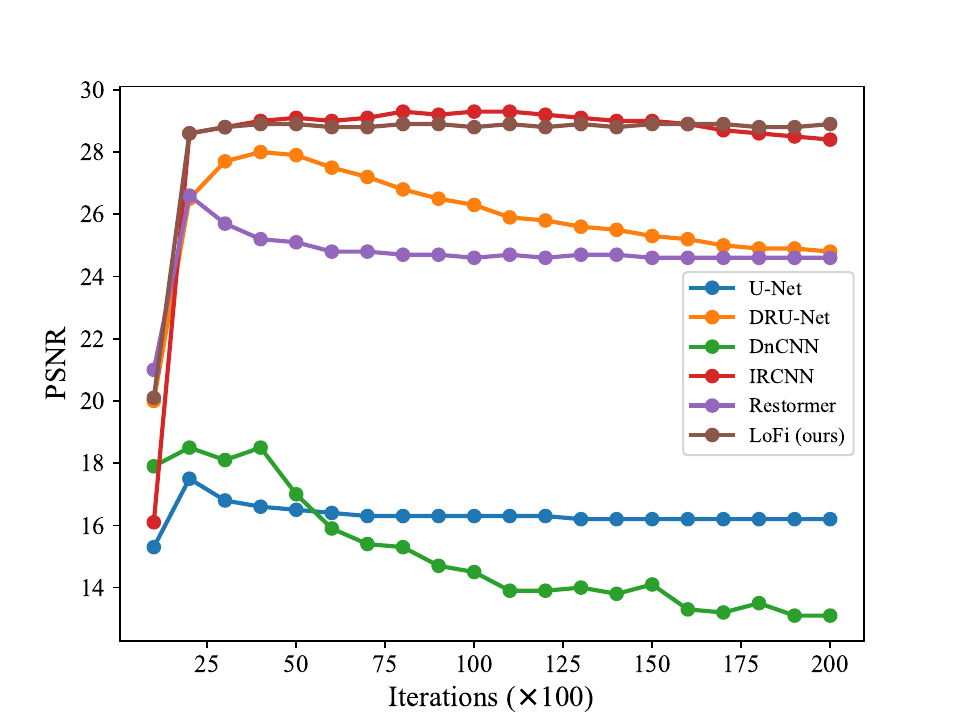}
	\caption{Comparative analysis for image denoising at resolution $512 \times 512$ where different models are trained on a tiny dataset with 9 training samples. The PSNR of the reconstructed test samples is demonstrated per iterations during training. CNNs, in particular multiscale versions, show severe overfitting while LoFi shows a robust convergence and significantly outperforms CNNs thanks to its locality design.}
	\label{fig: overfitting}
\end{figure}

Figure~\ref{fig: overfitting} presents the PSNR of the reconstructed test image across training iterations for each model. While CNNs initially perform well, they quickly overfit the training data, leading to a sharp drop in test image performance. Restormer and IRCNN demonstrate better generalization but also begin to overfit after 2500 and 10000 iterations, respectively. LoFi on the other hand generalizes strongly, avoiding overfitting even with 3M parameters and a tiny dataset. This is without any explicit regularization or early stopping. Final reconstructions after 20000 iterations are illustrated in Figure~\ref{fig: results_denoising_set10}. Please also see Section \ref{app: denoising_celeba} in the supplementary materials for additional experiments showcasing LoFi's performance when trained on a large-scale dataset (CelebA-HQ) with high-resolution images.

\begin{figure*}
	\centering
    \includegraphics[width=\linewidth]{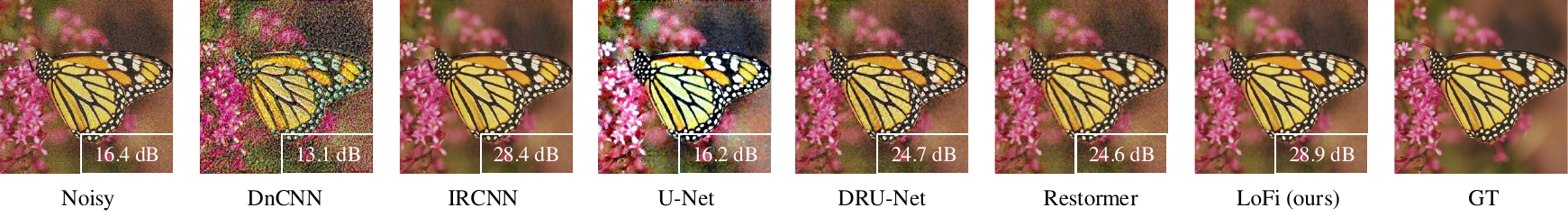}
	\caption{Performance comparison on image denoising where different models are trained for 20000 iterations on a tiny training set with 9 images shown in Figure~\ref{fig: training_set}.}
	\label{fig: results_denoising_set10}
\end{figure*}

\begin{figure*}[t]
    \centering
    \includegraphics[width = \textwidth]{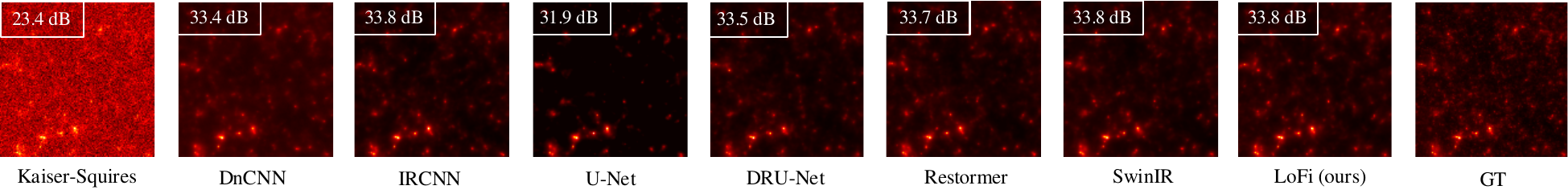}
    \\ [-5pt]
    \caption{Performance comparison on reconstructed convergence fields for dark matter mapping at resolution $128 \times 128$.}
    \label{fig: results_mass_kapp_128}
\end{figure*}

\begin{figure}[t]
    \centering
    \includegraphics[width = 0.45\textwidth]{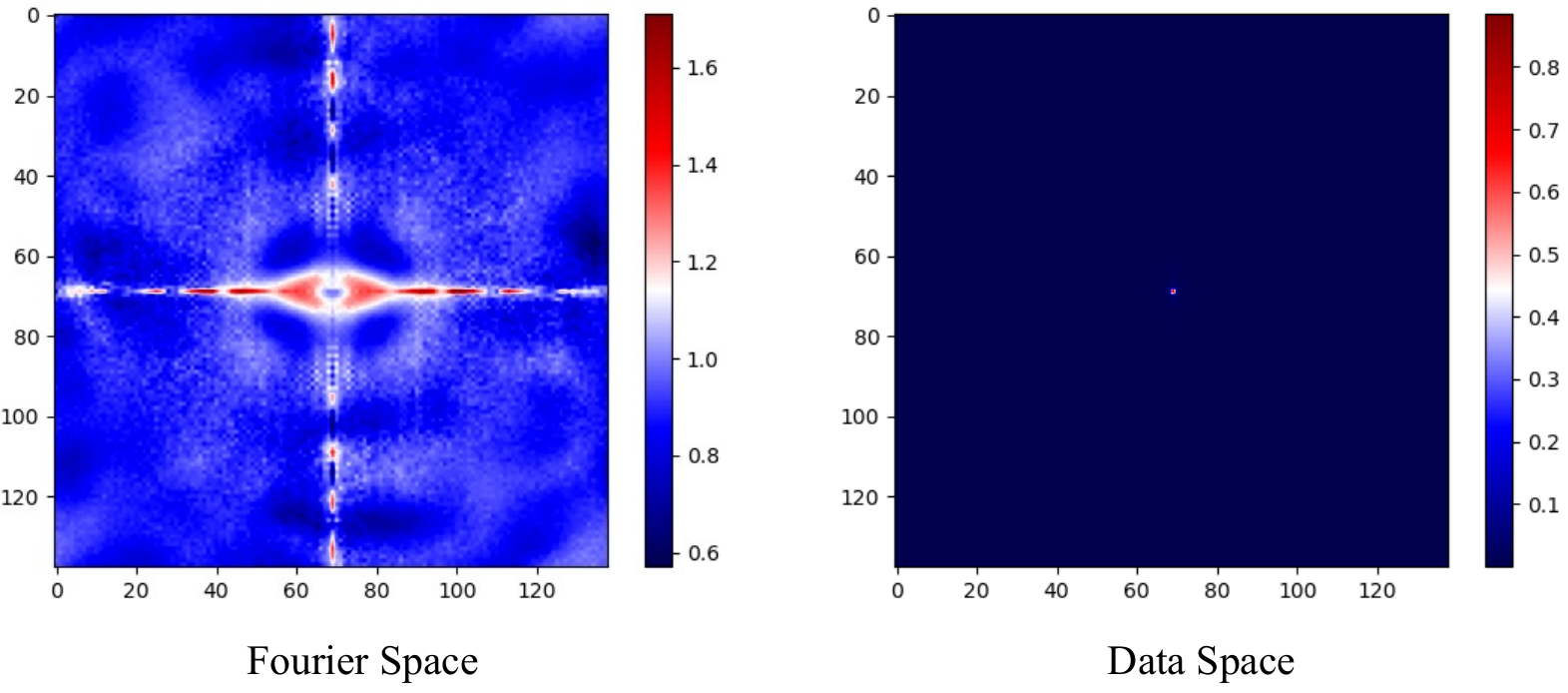}
    \\ [-5pt]
    \caption{The learned filter $\rvH$ in \eqref{eq: fourier filter}.}
    \label{fig: filter}
\end{figure}

\paragraph{Dark matter mapping}
We finally apply LoFi on the naive KS inversion $\kappa_\text{KS}$ to recover the convergence field $\kappa$. In Figure~\ref{fig: results_mass_kapp_128}, the reconstructed convergence fields from the KS image are presented for various models at resolution $128 \times 128$. The performance parity between LoFi and baselines, especially ViTs which can process long-range dependencies, indicates the validity of the locality property derived from \eqref{eq: KS}. In Figure \ref{fig: filter}, we also demonstrate the learned filter $\rvH$ in \eqref{eq: fourier filter}; this figure can be useful to understand how it processes input KS images. 

\subsection{Computational efficiency}
\label{sec: computation}

\begin{figure*}
\centering
\begin{subfigure}{0.5\textwidth}
  \centering
 \includegraphics[width=0.95\textwidth]{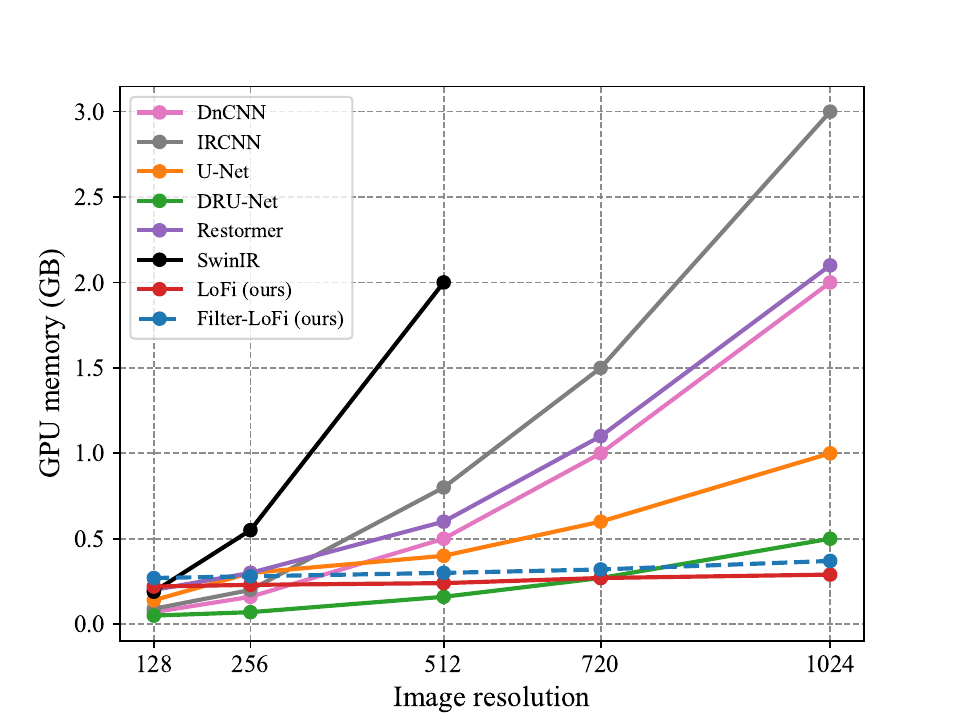}
\caption{Memory footprint}
\label{fig: gpu_test}
\end{subfigure}%
\begin{subfigure}{0.5\textwidth}
\centering
\includegraphics[width=0.95\textwidth]{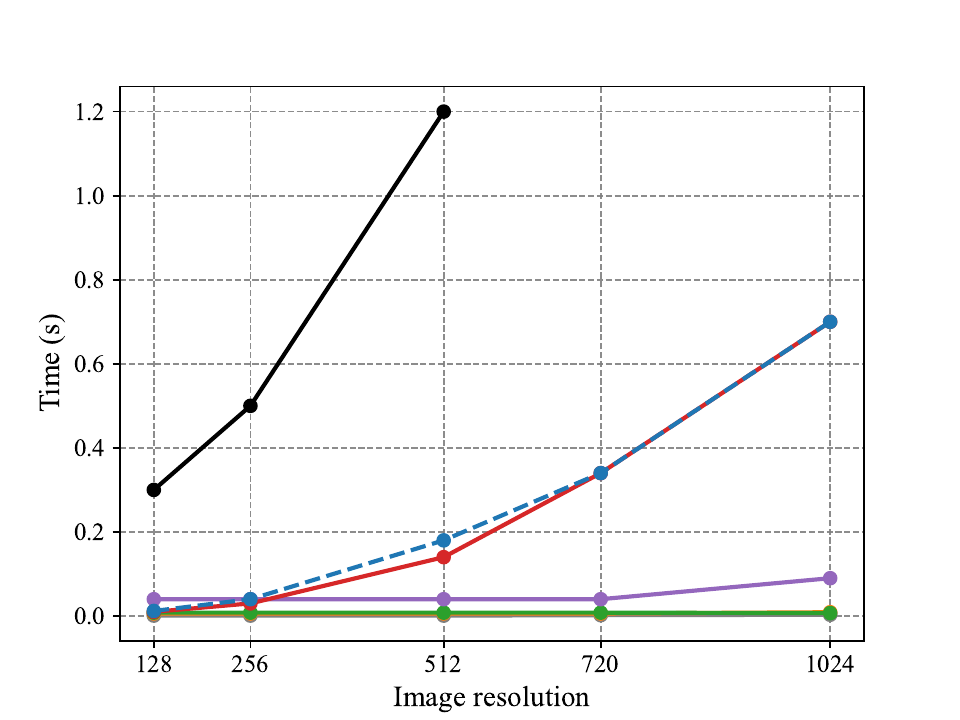}
\caption{Inference time}
\label{fig: time_test}
\end{subfigure}
\caption{The memory and time requirements during inference for different models}
\label{fig: inference computation}
\end{figure*}

As discussed in Section~\ref{sec: training}, LoFi can be trained and run on machines with small GPUs by reducing the pixel batch size in both training and inference. 
Figure \ref{fig: training computation} compares memory usage and training time for 100 iterations with batch size 1 across different models and image resolutions. In this experiment, we used pixel batch size 512 for LoFi during training. This analysis shows that while the required memory and training time for CNNs and ViTs scales with image resolution, LoFi's memory usage and training time remains nearly constant enabling efficient training on high-resolution images.

Moreover, LoFi training is fast even for large images. For example, training on $1024 \times 1024$ images using 29900 samples from the CelebA-HQ dataset was completed in only 8 hours (200 epochs) on a single A100 GPU, requiring only 15 GB memory with object and pixel batch sizes 64 and 512 respectively. Additional details and qualitative results for this experiment can be found in Section \ref{app: denoising_celeba}.

While LoFi is efficient during training, inference on large images can become expensive due to coordinate-based image synthesis. This requires iteratively reconstructing large images, as generating all pixels at once would need a bigger memory. Figure \ref{fig: inference computation} compares the memory usage and inference time of different architectures. In this experiment, LoFi generates 16384 pixels per iteration. We mention potential strategies to improve inference efficiency in Section \ref{sec: conclusion}.

\subsection{Reconstruction analysis and interpretation}
\label{sec: adaptive patch experiment}
\begin{figure}
	\centering
    \includegraphics[width=0.9\linewidth]{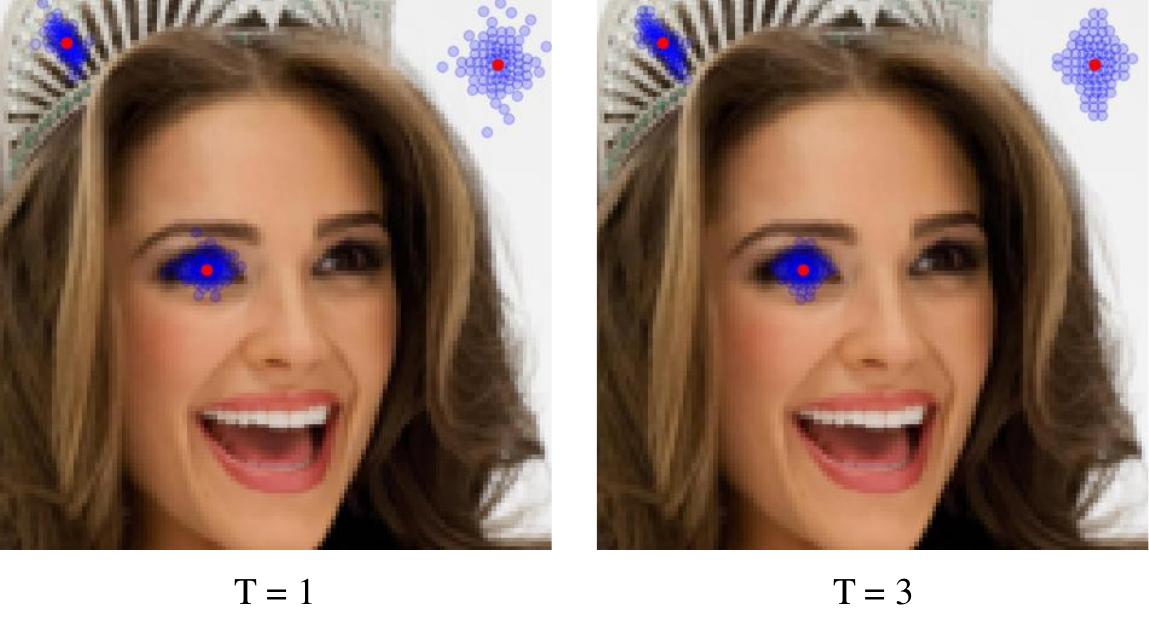}
	\caption{The learned patch geometry for different pixels. As expected, larger $T$ results in more accurate localization of the relevant information. The red dots indicate the query pixels, while the blue dots represent the estimated positions of relevant features to be extracted from the noisy image.}
	\label{fig: patch geometry}
\end{figure}

\begin{figure*}
	\centering
    \includegraphics[width=0.8\linewidth]{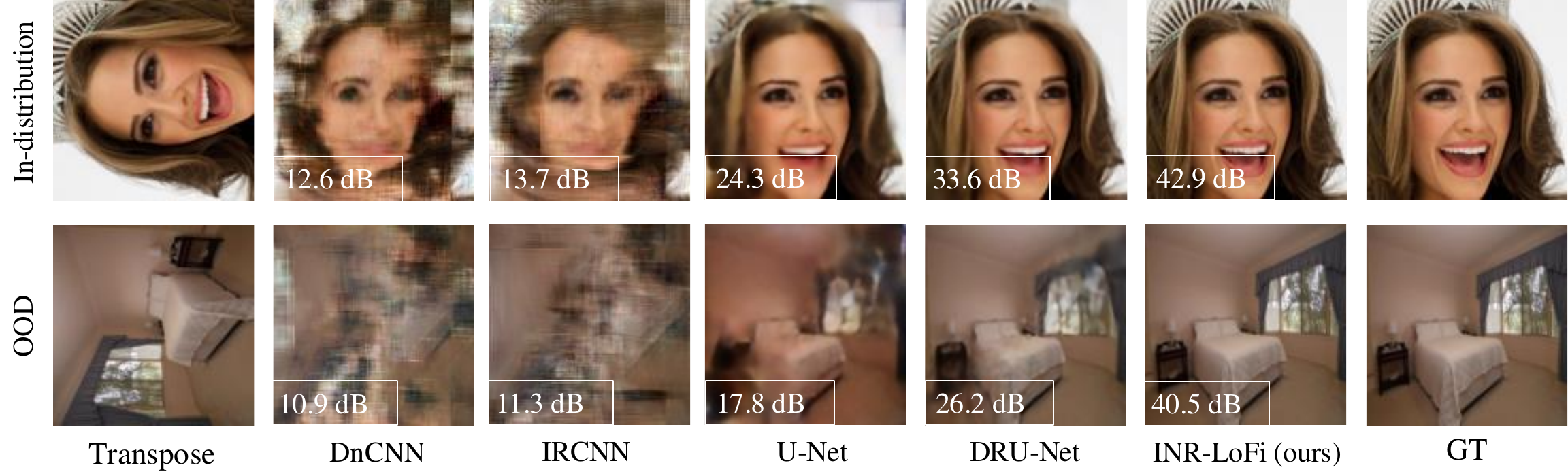}
	\caption{Performance comparison on toy experiment of recovering an image from its transpose.}
	\label{fig: transpose}
\end{figure*}
As discussed in Section~\ref{sec: Coordinate-based patch geometry}, LoFi adaptively estimates patch geometry for each pixel based on local information. Figure~\ref{fig: patch geometry} illustrates the learned patch geometry across different pixels for image denoising, where the estimated positions of relevant features align with intuitive expectations. This ability to pinpoint features contributing to each pixel’s estimation can be valuable for interpretation. Additionally, the figure shows that more iterations (larger values of $T$) result in a more precise localization of relevant information.

\subsection{Learning the imaging geometry with INR-LoFi}
\label{sec: exp_INR_LoFi}

In this section, we study the performance of different models on a toy experiment where we have changes in the coordinate system. We consider recovering an image from its transpose; a simple task where we only have changes in the coordinate system while the image content remains unchanged. We train LoFi and CNNs on 29900 training samples from CelebA-HQ~\cite{karras2017progressive} dataset. Figure~\ref{fig: transpose} showcases the performance of different models on both in-distribution (CelebA-HQ test samples) and OOD data (LSUN-bedroom \cite{yu2015lsun}). Despite the simplicity of the task and access to a large-scale dataset, CNNs show poor performance as they can only learn the changes in the image content while neglecting the changes in the coordinate system. Unlike CNNs, our proposed INR-LoFi can conveniently learn these changes in the coordinate system resulting in significantly better reconstructions.

\subsection{Low-resolution priors for high-resolution image reconstruction with LoFi-ADMM}
\label{sec: pnp_experiment}

As discussed in Section~\ref{sec: LoFi-admm}, LoFi can be used as a low-dimensional prior for processing the measurements acquired at higher resolutions. To this end, we train LoFi denoiser on low-resolution images and use it as a prior in LoFi-ADMM for solving two inverse problems at higher resolutions; image in-painting and radio interferometric imaging. 

\paragraph{Image inpainting}

\begin{figure}
    \centering
    \includegraphics[width = 0.45\textwidth]{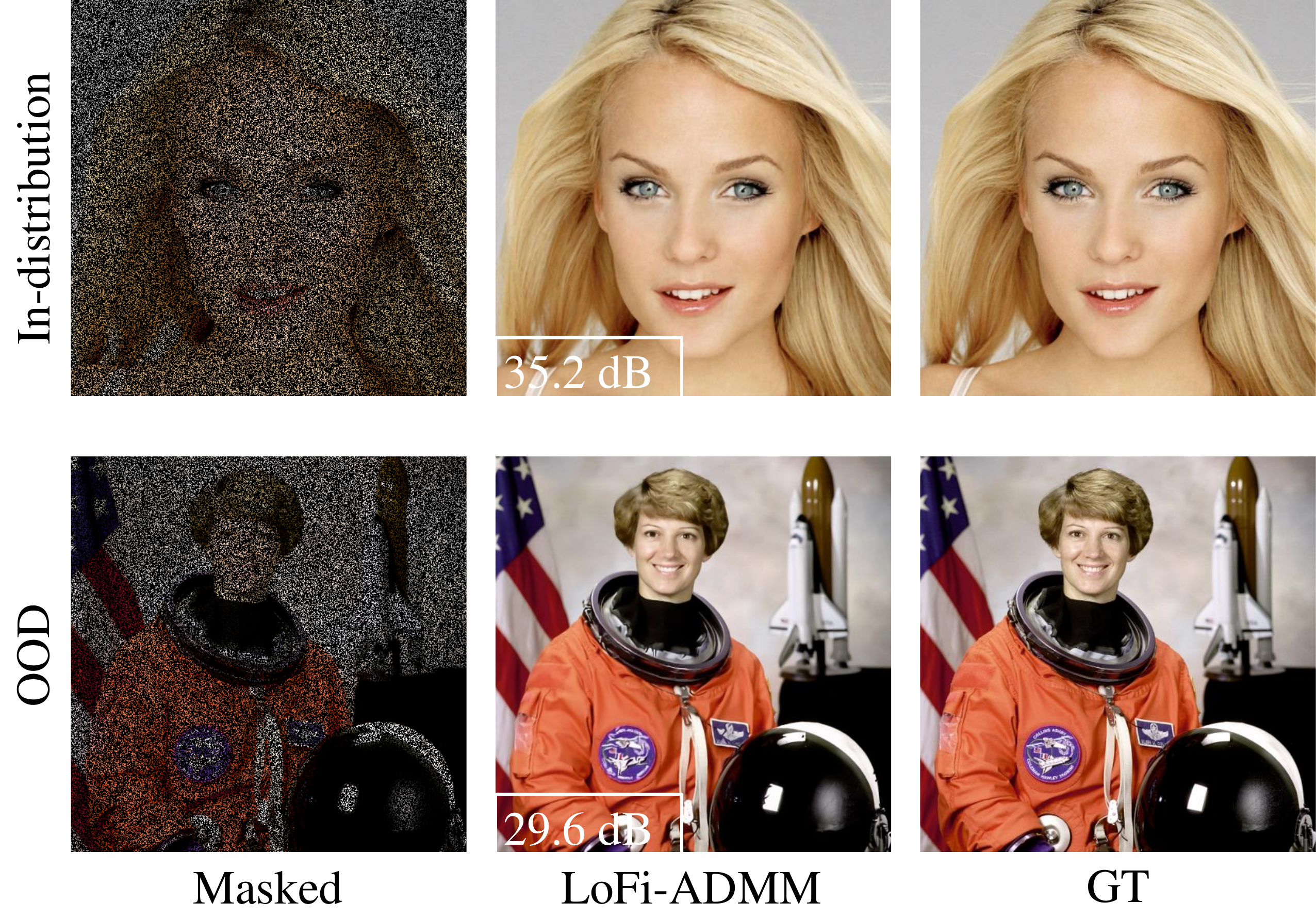}
    \\[-5pt]
    \caption{LoFi-ADMM performance on image in-painting ($p = 30\%$) for in-distribution and OOD data at resolution $512 \times 512$ when LoFi denoiser is trained on CelebA-HQ samples in low resolution $128 \times 128$.}
    \label{fig: exp_inpainting}
\end{figure}

We train LoFi denoiser on 29900 samples from the CelebA-HQ~\cite{karras2017progressive} in low resolution $128 \times 128$ where we consider zero-mean Gaussian noise $\rvn$ with $\sigma = 0.15$ and normalize images between 0 and 1. In LoFi-ADMM, we set $\alpha = 0.05$ and execute it for 90 iterations, the hyper-parameters we found suitable to ensure convergence.
Now, we run LoFi-ADMM to solve the image-inpainting problem at a higher resolution $512 \times 512$ where $p = 30\%$ of the image pixels are randomly masked. Figure~\ref{fig: exp_inpainting} showcases the performance of the LoFi-ADMM on both in-distribution and OOD data. This experiment shows that once LoFi denoiser is trained, it can be used for solving general inverse problems at any higher resolution with a strong generalization.

\begin{figure}
    \centering
    \includegraphics[width = 0.45\textwidth]{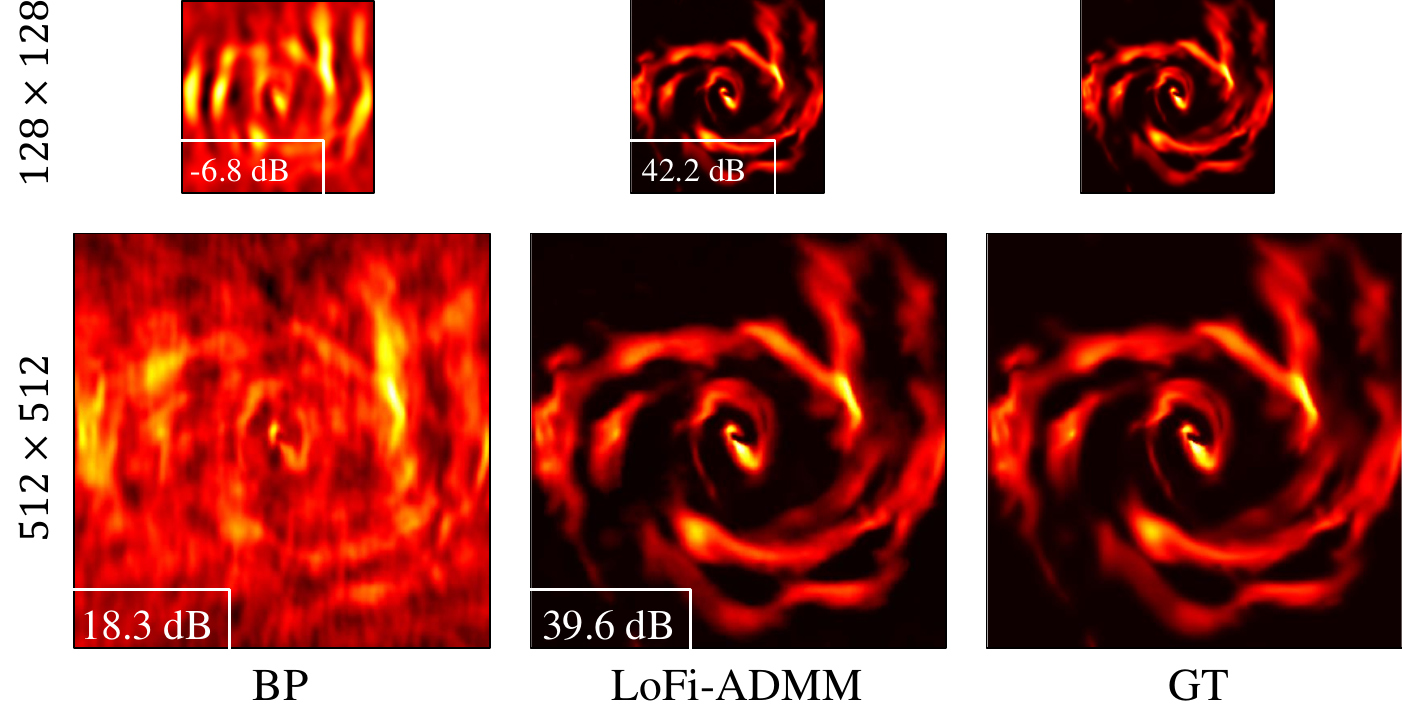}
    \\ [-5pt]
    \caption{LoFi-ADMM performance on radio interferometry; the LoFi prior is trained on low-resolution images $128 \times 128$.}
    \label{fig: exp_radio}
\end{figure}

\paragraph{Radio interferometry}
We next apply LoFi-ADMM to radio interferometric imaging, please refer to Section~\ref{sec: radio} in the supplementary material for further information regarding the forward model. We use non-uniform fast Fourier transform (NUFFT) to simulate radio interferometric measurements from $128 \times 128$ images using an upsampling factor of 2 and Kaiser--Bessel kernels with
a size of 6 $\times$ 6 pixels; we use the simulated uv-coverage of the MeerKAT telescope and the measurements with complex Gaussian noise at the SNR of 30dB.
We train LoFi denoiser on 13000 images of simulated galaxies created from
the IllustrisTNG simulations \cite{nelson2019first}. Figure~\ref{fig: exp_radio} illustrates the performance of the LoFi-ADMM at original and higher resolutions. This experiment shows that LoFi-ADMM can conveniently exploit the measurements obtained from higher-resolution images for more accurate reconstructions with more details.

\section{Limitations and conclusion}
\label{sec: conclusion}

We introduced LoFi, a scalable local image reconstruction framework that stands out for its strong generalization on OOD data and resolution agnostic memory usage during training. LoFi achieves strong performance on local inverse problems such as LDCT and image denoising, all while maintaining a significantly smaller memory footprint compared to standard CNNs and vision transformers. The favorable inductive bias of LoFi allows us to train it on very small datasets without overfitting or explicit regularization. Our experiments further demonstrate the effectiveness of LoFi as a denoising prior within the PnP framework, showcasing its ability to reconstruct images at arbitrary resolutions for tasks such as image in-painting and radio interferometric imaging.

Although LoFi's pixel-based pipeline significantly reduces memory requirements during training, it comes with trade-offs. Like INRs, the computational cost of LoFi in inference increases quadratically with image resolution as shown in Figure \ref{fig: time_test}. Recent studies \cite{he2023dynamic, he2024latent} have proposed strategies to enhance the efficiency of continuous image representation in INRs by increasing shared computations across pixels, thereby reducing computational complexity. Implementing similar strategies in LoFi could potentially reduce computations during inference, which we leave as a direction for future work.

\newpage
\bibliography{main}

\begin{thebibliography}{100}

\bibitem{wang2008outlook}
G.~Wang, H.~Yu, and B.~De~Man, ``An outlook on x-ray ct research and development,'' {\em Medical physics}, vol.~35, no.~3, pp.~1051--1064, 2008.

\bibitem{holler2017high}
M.~Holler, M.~Guizar-Sicairos, E.~H. Tsai, R.~Dinapoli, E.~M{\"u}ller, O.~Bunk, J.~Raabe, and G.~Aeppli, ``High-resolution non-destructive three-dimensional imaging of integrated circuits,'' {\em Nature}, vol.~543, no.~7645, pp.~402--406, 2017.

\bibitem{blahut2004theory}
R.~E. Blahut, {\em Theory of remote image formation}.
\newblock Cambridge University Press, 2004.

\bibitem{kaiser1993}
N.~{Kaiser} and G.~{Squires}, ``{Mapping the Dark Matter with Weak Gravitational Lensing},'' {\em ApJ}, vol.~404, p.~441, Feb. 1993.

\bibitem{ronneberger2015u}
O.~Ronneberger, P.~Fischer, and T.~Brox, ``U-net: Convolutional networks for biomedical image segmentation,'' in {\em Medical Image Computing and Computer-Assisted Intervention--MICCAI 2015: 18th International Conference, Munich, Germany, October 5-9, 2015, Proceedings, Part III 18}, pp.~234--241, Springer, 2015.

\bibitem{jin2017deep}
K.~H. Jin, M.~T. McCann, E.~Froustey, and M.~Unser, ``Deep convolutional neural network for inverse problems in imaging,'' {\em IEEE transactions on image processing}, vol.~26, no.~9, pp.~4509--4522, 2017.

\bibitem{hyun2018deep}
C.~M. Hyun, H.~P. Kim, S.~M. Lee, S.~Lee, and J.~K. Seo, ``Deep learning for undersampled mri reconstruction,'' {\em Physics in Medicine \& Biology}, vol.~63, no.~13, p.~135007, 2018.

\bibitem{mars2024learned}
M.~Mars, M.~M. Betcke, and J.~D. McEwen, ``Learned radio interferometric imaging for varying visibility coverage,'' {\em arXiv preprint arXiv:2405.08958}, 2024.

\bibitem{jeffrey2020deep}
N.~Jeffrey, F.~Lanusse, O.~Lahav, and J.-L. Starck, ``Deep learning dark matter map reconstructions from des sv weak lensing data,'' {\em Monthly Notices of the Royal Astronomical Society}, vol.~492, no.~4, pp.~5023--5029, 2020.

\bibitem{liu2022learning}
T.~Liu, A.~Chaman, D.~Belius, and I.~Dokmani{\'c}, ``Learning multiscale convolutional dictionaries for image reconstruction,'' {\em IEEE Transactions on Computational Imaging}, vol.~8, pp.~425--437, 2022.

\bibitem{zhang2017beyond}
K.~Zhang, W.~Zuo, Y.~Chen, D.~Meng, and L.~Zhang, ``Beyond a gaussian denoiser: Residual learning of deep cnn for image denoising,'' {\em IEEE transactions on image processing}, vol.~26, no.~7, pp.~3142--3155, 2017.

\bibitem{liang2021swinir}
J.~Liang, J.~Cao, G.~Sun, K.~Zhang, L.~Van~Gool, and R.~Timofte, ``Swinir: Image restoration using swin transformer,'' in {\em Proceedings of the IEEE/CVF international conference on computer vision}, pp.~1833--1844, 2021.

\bibitem{wang2021pyramid}
W.~Wang, E.~Xie, X.~Li, D.-P. Fan, K.~Song, D.~Liang, T.~Lu, P.~Luo, and L.~Shao, ``Pyramid vision transformer: A versatile backbone for dense prediction without convolutions,'' in {\em Proceedings of the IEEE/CVF international conference on computer vision}, pp.~568--578, 2021.

\bibitem{sepehri2024serpent}
M.~S. Sepehri, Z.~Fabian, and M.~Soltanolkotabi, ``Serpent: Scalable and efficient image restoration via multi-scale structured state space models,'' {\em arXiv preprint arXiv:2403.17902}, 2024.

\bibitem{zhang2021plug}
K.~Zhang, Y.~Li, W.~Zuo, L.~Zhang, L.~Van~Gool, and R.~Timofte, ``Plug-and-play image restoration with deep denoiser prior,'' {\em IEEE Transactions on Pattern Analysis and Machine Intelligence}, vol.~44, no.~10, pp.~6360--6376, 2021.

\bibitem{fabian2022humus}
Z.~Fabian, B.~Tinaz, and M.~Soltanolkotabi, ``Humus-net: Hybrid unrolled multi-scale network architecture for accelerated mri reconstruction,'' {\em Advances in Neural Information Processing Systems}, vol.~35, pp.~25306--25319, 2022.

\bibitem{aggarwal2018modl}
H.~K. Aggarwal, M.~P. Mani, and M.~Jacob, ``Modl: Model-based deep learning architecture for inverse problems,'' {\em IEEE transactions on medical imaging}, vol.~38, no.~2, pp.~394--405, 2018.

\bibitem{sitzmann2020implicit}
V.~Sitzmann, J.~Martel, A.~Bergman, D.~Lindell, and G.~Wetzstein, ``Implicit neural representations with periodic activation functions,'' {\em Advances in neural information processing systems}, vol.~33, pp.~7462--7473, 2020.

\bibitem{atzmon2020sal}
M.~Atzmon and Y.~Lipman, ``Sal: Sign agnostic learning of shapes from raw data,'' in {\em Proceedings of the IEEE/CVF conference on computer vision and pattern recognition}, pp.~2565--2574, 2020.

\bibitem{chabra2020deep}
R.~Chabra, J.~E. Lenssen, E.~Ilg, T.~Schmidt, J.~Straub, S.~Lovegrove, and R.~Newcombe, ``Deep local shapes: Learning local sdf priors for detailed 3d reconstruction,'' in {\em Computer Vision--ECCV 2020: 16th European Conference, Glasgow, UK, August 23--28, 2020, Proceedings, Part XXIX 16}, pp.~608--625, Springer, 2020.

\bibitem{chen2019learning}
Z.~Chen and H.~Zhang, ``Learning implicit fields for generative shape modeling,'' in {\em Proceedings of the IEEE/CVF conference on computer vision and pattern recognition}, pp.~5939--5948, 2019.

\bibitem{peng2020convolutional}
S.~Peng, M.~Niemeyer, L.~Mescheder, M.~Pollefeys, and A.~Geiger, ``Convolutional occupancy networks,'' in {\em Computer Vision--ECCV 2020: 16th European Conference, Glasgow, UK, August 23--28, 2020, Proceedings, Part III 16}, pp.~523--540, Springer, 2020.

\bibitem{jiang2020local}
C.~Jiang, A.~Sud, A.~Makadia, J.~Huang, M.~Nie{\ss}ner, T.~Funkhouser, {\em et~al.}, ``Local implicit grid representations for 3d scenes,'' in {\em Proceedings of the IEEE/CVF Conference on Computer Vision and Pattern Recognition}, pp.~6001--6010, 2020.

\bibitem{dupont2022data}
E.~Dupont, H.~Kim, S.~Eslami, D.~Rezende, and D.~Rosenbaum, ``From data to functa: Your data point is a function and you can treat it like one,'' {\em arXiv preprint arXiv:2201.12204}, 2022.

\bibitem{dupont2021generative}
E.~Dupont, Y.~W. Teh, and A.~Doucet, ``Generative models as distributions of functions,'' {\em arXiv preprint arXiv:2102.04776}, 2021.

\bibitem{susmelj2024uncertainty}
A.~Susmelj, M.~Macuglia, N.~Tagasovska, R.~Sutter, S.~Caprara, J.-P. Thiran, and E.~Konukoglu, ``Uncertainty modeling for fine-tuned implicit functions,'' {\em arXiv preprint arXiv:2406.12082}, 2024.

\bibitem{mildenhall2021nerf}
B.~Mildenhall, P.~P. Srinivasan, M.~Tancik, J.~T. Barron, R.~Ramamoorthi, and R.~Ng, ``Nerf: Representing scenes as neural radiance fields for view synthesis,'' {\em Communications of the ACM}, vol.~65, no.~1, pp.~99--106, 2021.

\bibitem{mohan2019robust}
S.~Mohan, Z.~Kadkhodaie, E.~P. Simoncelli, and C.~Fernandez-Granda, ``Robust and interpretable blind image denoising via bias-free convolutional neural networks,'' {\em ICLR}, 2020.

\bibitem{wang2022uformer}
Z.~Wang, X.~Cun, J.~Bao, W.~Zhou, J.~Liu, and H.~Li, ``Uformer: A general u-shaped transformer for image restoration,'' in {\em Proceedings of the IEEE/CVF conference on computer vision and pattern recognition}, pp.~17683--17693, 2022.

\bibitem{zhao2023comprehensive}
H.~Zhao, Y.~Gou, B.~Li, D.~Peng, J.~Lv, and X.~Peng, ``Comprehensive and delicate: An efficient transformer for image restoration,'' in {\em Proceedings of the IEEE/CVF conference on computer vision and pattern recognition}, pp.~14122--14132, 2023.

\bibitem{osher2005iterative}
S.~Osher, M.~Burger, D.~Goldfarb, J.~Xu, and W.~Yin, ``An iterative regularization method for total variation-based image restoration,'' {\em Multiscale Modeling \& Simulation}, vol.~4, no.~2, pp.~460--489, 2005.

\bibitem{kothari2021trumpets}
K.~Kothari, A.~Khorashadizadeh, M.~de~Hoop, and I.~Dokmani{\'c}, ``Trumpets: Injective flows for inference and inverse problems,'' in {\em Uncertainty in Artificial Intelligence}, pp.~1269--1278, PMLR, 2021.

\bibitem{jalal2021robust}
A.~Jalal, M.~Arvinte, G.~Daras, E.~Price, A.~G. Dimakis, and J.~Tamir, ``Robust compressed sensing mri with deep generative priors,'' {\em Advances in Neural Information Processing Systems}, vol.~34, pp.~14938--14954, 2021.

\bibitem{chung2022diffusion}
H.~Chung, J.~Kim, M.~T. Mccann, M.~L. Klasky, and J.~C. Ye, ``Diffusion posterior sampling for general noisy inverse problems,'' {\em arXiv preprint arXiv:2209.14687}, 2022.

\bibitem{hu2023restoration}
Y.~Hu, M.~Delbracio, P.~Milanfar, and U.~S. Kamilov, ``A restoration network as an implicit prior,'' {\em ICLR}, 2024.

\bibitem{levin2011natural}
A.~Levin and B.~Nadler, ``Natural image denoising: Optimality and inherent bounds,'' in {\em CVPR 2011}, pp.~2833--2840, IEEE, 2011.

\bibitem{mayo1997simulated}
J.~R. Mayo, K.~P. Whittall, A.~N. Leung, T.~E. Hartman, C.~S. Park, S.~L. Primack, G.~K. Chambers, M.~K. Limkeman, T.~L. Toth, and S.~H. Fox, ``Simulated dose reduction in conventional chest ct: validation study.,'' {\em Radiology}, vol.~202, no.~2, pp.~453--457, 1997.

\bibitem{brenner2007computed}
D.~J. Brenner and E.~J. Hall, ``Computed tomography—an increasing source of radiation exposure,'' {\em New England journal of medicine}, vol.~357, no.~22, pp.~2277--2284, 2007.

\bibitem{einstein2007estimating}
A.~J. Einstein, M.~J. Henzlova, and S.~Rajagopalan, ``Estimating risk of cancer associated with radiation exposure from 64-slice computed tomography coronary angiography,'' {\em Jama}, vol.~298, no.~3, pp.~317--323, 2007.

\bibitem{helgason1980radon}
S.~Helgason and S.~Helgason, {\em The radon transform}, vol.~2.
\newblock Springer, 1980.

\bibitem{kiss2024learned}
M.~B. Kiss, A.~Biguri, C.-B. Sch{\"o}nlieb, K.~J. Batenburg, and F.~Lucka, ``Learned denoising with simulated and experimental low-dose ct data,'' {\em arXiv preprint arXiv:2408.08115}, 2024.

\bibitem{mandelbaum2018}
R.~Mandelbaum, ``Weak lensing for precision cosmology,'' {\em Annual Review of Astronomy and Astrophysics}, vol.~56, no.~1, pp.~393--433, 2018.

\bibitem{burger2012image}
H.~C. Burger, C.~J. Schuler, and S.~Harmeling, ``Image denoising: Can plain neural networks compete with bm3d?,'' in {\em 2012 IEEE conference on computer vision and pattern recognition}, pp.~2392--2399, IEEE, 2012.

\bibitem{zoran2011learning}
D.~Zoran and Y.~Weiss, ``From learning models of natural image patches to whole image restoration,'' in {\em 2011 international conference on computer vision}, pp.~479--486, IEEE, 2011.

\bibitem{zhang2017learning}
K.~Zhang, W.~Zuo, S.~Gu, and L.~Zhang, ``Learning deep cnn denoiser prior for image restoration,'' in {\em Proceedings of the IEEE conference on computer vision and pattern recognition}, pp.~3929--3938, 2017.

\bibitem{liu2018multi}
P.~Liu, H.~Zhang, K.~Zhang, L.~Lin, and W.~Zuo, ``Multi-level wavelet-cnn for image restoration,'' in {\em Proceedings of the IEEE conference on computer vision and pattern recognition workshops}, pp.~773--782, 2018.

\bibitem{park2019densely}
B.~Park, S.~Yu, and J.~Jeong, ``Densely connected hierarchical network for image denoising,'' in {\em Proceedings of the IEEE/CVF conference on computer vision and pattern recognition workshops}, pp.~0--0, 2019.

\bibitem{jia2021ddunet}
F.~Jia, W.~H. Wong, and T.~Zeng, ``Ddunet: Dense dense u-net with applications in image denoising,'' in {\em Proceedings of the IEEE/CVF international conference on computer vision}, pp.~354--364, 2021.

\bibitem{he2016deep}
K.~He, X.~Zhang, S.~Ren, and J.~Sun, ``Deep residual learning for image recognition,'' in {\em Proceedings of the IEEE conference on computer vision and pattern recognition}, pp.~770--778, 2016.

\bibitem{dosovitskiy2020image}
A.~Dosovitskiy, L.~Beyer, A.~Kolesnikov, D.~Weissenborn, X.~Zhai, T.~Unterthiner, M.~Dehghani, M.~Minderer, G.~Heigold, S.~Gelly, {\em et~al.}, ``An image is worth 16x16 words: Transformers for image recognition at scale,'' {\em arXiv preprint arXiv:2010.11929}, 2020.

\bibitem{chen2021pre}
H.~Chen, Y.~Wang, T.~Guo, C.~Xu, Y.~Deng, Z.~Liu, S.~Ma, C.~Xu, C.~Xu, and W.~Gao, ``Pre-trained image processing transformer,'' in {\em Proceedings of the IEEE/CVF conference on computer vision and pattern recognition}, pp.~12299--12310, 2021.

\bibitem{mansour2022image}
Y.~Mansour, K.~Lin, and R.~Heckel, ``Image-to-image mlp-mixer for image reconstruction,'' {\em arXiv preprint arXiv:2202.02018}, 2022.

\bibitem{zamir2022restormer}
S.~W. Zamir, A.~Arora, S.~Khan, M.~Hayat, F.~S. Khan, and M.-H. Yang, ``Restormer: Efficient transformer for high-resolution image restoration,'' in {\em Proceedings of the IEEE/CVF conference on computer vision and pattern recognition}, pp.~5728--5739, 2022.

\bibitem{heo2021rethinking}
B.~Heo, S.~Yun, D.~Han, S.~Chun, J.~Choe, and S.~J. Oh, ``Rethinking spatial dimensions of vision transformers,'' in {\em Proceedings of the IEEE/CVF international conference on computer vision}, pp.~11936--11945, 2021.

\bibitem{vlavsic2022implicit}
T.~Vla{\v{s}}i{\'c}, H.~Nguyen, A.~Khorashadizadeh, and I.~Dokmani{\'c}, ``Implicit neural representation for mesh-free inverse obstacle scattering,'' in {\em 2022 56th Asilomar Conference on Signals, Systems, and Computers}, pp.~947--952, IEEE, 2022.

\bibitem{chen2021learning}
Y.~Chen, S.~Liu, and X.~Wang, ``Learning continuous image representation with local implicit image function,'' in {\em Proceedings of the IEEE/CVF conference on computer vision and pattern recognition}, pp.~8628--8638, 2021.

\bibitem{khorashadizadeh2022funknn}
A.~Khorashadizadeh, A.~Chaman, V.~Debarnot, and I.~Dokmani{\'c}, ``Funknn: Neural interpolation for functional generation,'' {\em ICLR}, 2023.

\bibitem{buades2005non}
A.~Buades, B.~Coll, and J.-M. Morel, ``A non-local algorithm for image denoising,'' in {\em 2005 IEEE computer society conference on computer vision and pattern recognition (CVPR'05)}, vol.~2, pp.~60--65, Ieee, 2005.

\bibitem{dabov2007image}
K.~Dabov, A.~Foi, V.~Katkovnik, and K.~Egiazarian, ``Image denoising by sparse 3-d transform-domain collaborative filtering,'' {\em IEEE Transactions on image processing}, vol.~16, no.~8, pp.~2080--2095, 2007.

\bibitem{tolstikhin2021mlp}
I.~O. Tolstikhin, N.~Houlsby, A.~Kolesnikov, L.~Beyer, X.~Zhai, T.~Unterthiner, J.~Yung, A.~Steiner, D.~Keysers, J.~Uszkoreit, {\em et~al.}, ``Mlp-mixer: An all-mlp architecture for vision,'' {\em Advances in neural information processing systems}, vol.~34, pp.~24261--24272, 2021.

\bibitem{trockman2022patches}
A.~Trockman and J.~Z. Kolter, ``Patches are all you need?,'' {\em arXiv preprint arXiv:2201.09792}, 2022.

\bibitem{ding2023patched}
Z.~Ding, M.~Zhang, J.~Wu, and Z.~Tu, ``Patched denoising diffusion models for high-resolution image synthesis,'' in {\em ICLR}, 2024.

\bibitem{wang2024patch}
Z.~Wang, Y.~Jiang, H.~Zheng, P.~Wang, P.~He, Z.~Wang, W.~Chen, M.~Zhou, {\em et~al.}, ``Patch diffusion: Faster and more data-efficient training of diffusion models,'' {\em Advances in Neural Information Processing Systems}, vol.~36, 2024.

\bibitem{gilton2019learned}
D.~Gilton, G.~Ongie, and R.~Willett, ``Learned patch-based regularization for inverse problems in imaging,'' in {\em 2019 ieee 8th international workshop on computational advances in multi-sensor adaptive processing (camsap)}, pp.~211--215, IEEE, 2019.

\bibitem{altekruger2022patchnr}
F.~Altekr{\"u}ger, A.~Denker, P.~Hagemann, J.~Hertrich, P.~Maass, and G.~Steidl, ``Patchnr: Learning from small data by patch normalizing flow regularization,'' {\em arXiv preprint arXiv:2205.12021}, 2022.

\bibitem{hertrich2022wasserstein}
J.~Hertrich, A.~Houdard, and C.~Redenbach, ``Wasserstein patch prior for image superresolution,'' {\em IEEE Transactions on Computational Imaging}, vol.~8, pp.~693--704, 2022.

\bibitem{piening2024learning}
M.~Piening, F.~Altekr{\"u}ger, J.~Hertrich, P.~Hagemann, A.~Walther, and G.~Steidl, ``Learning from small data sets: Patch-based regularizers in inverse problems for image reconstruction,'' {\em GAMM-Mitteilungen}, p.~e202470002, 2024.

\bibitem{dai2017deformable}
J.~Dai, H.~Qi, Y.~Xiong, Y.~Li, G.~Zhang, H.~Hu, and Y.~Wei, ``Deformable convolutional networks,'' in {\em Proceedings of the IEEE international conference on computer vision}, pp.~764--773, 2017.

\bibitem{diwakar2018}
M.~Diwakar and M.~Kumar, ``A review on ct image noise and its denoising,'' {\em Biomedical Signal Processing and Control}, vol.~42, pp.~73--88, 2018.

\bibitem{kothari2020learning}
K.~Kothari, M.~de~Hoop, and I.~Dokmani{\'c}, ``Learning the geometry of wave-based imaging,'' {\em Advances in Neural Information Processing Systems}, vol.~33, pp.~8318--8329, 2020.

\bibitem{mccann2017convolutional}
M.~T. McCann, K.~H. Jin, and M.~Unser, ``Convolutional neural networks for inverse problems in imaging: A review,'' {\em IEEE Signal Processing Magazine}, vol.~34, no.~6, pp.~85--95, 2017.

\bibitem{wei2018deep}
Z.~Wei and X.~Chen, ``Deep-learning schemes for full-wave nonlinear inverse scattering problems,'' {\em IEEE Transactions on Geoscience and Remote Sensing}, vol.~57, no.~4, pp.~1849--1860, 2018.

\bibitem{khorashadizadeh2023conditional}
A.~Khorashadizadeh, K.~Kothari, L.~Salsi, A.~A. Harandi, M.~de~Hoop, and I.~Dokmani{\'c}, ``Conditional injective flows for bayesian imaging,'' {\em IEEE Transactions on Computational Imaging}, vol.~9, pp.~224--237, 2023.

\bibitem{bora2017compressed}
A.~Bora, A.~Jalal, E.~Price, and A.~G. Dimakis, ``Compressed sensing using generative models,'' in {\em International conference on machine learning}, pp.~537--546, PMLR, 2017.

\bibitem{kawar2022denoising}
B.~Kawar, M.~Elad, S.~Ermon, and J.~Song, ``Denoising diffusion restoration models,'' {\em Advances in Neural Information Processing Systems}, vol.~35, pp.~23593--23606, 2022.

\bibitem{khorashadizadeh2023deep}
A.~Khorashadizadeh, V.~Khorashadizadeh, S.~Eskandari, G.~A. Vandenbosch, and I.~Dokmani{\'c}, ``Deep injective prior for inverse scattering,'' {\em IEEE Transactions on Antennas and Propagation}, 2023.

\bibitem{liu2023optimization}
T.~Liu, T.~Yang, Q.~Zhang, and Q.~Lei, ``Optimization for amortized inverse problems,'' in {\em International Conference on Machine Learning}, pp.~22289--22319, PMLR, 2023.

\bibitem{venkatakrishnan2013plug}
S.~V. Venkatakrishnan, C.~A. Bouman, and B.~Wohlberg, ``Plug-and-play priors for model based reconstruction,'' in {\em 2013 IEEE global conference on signal and information processing}, pp.~945--948, IEEE, 2013.

\bibitem{chan2016plug}
S.~H. Chan, X.~Wang, and O.~A. Elgendy, ``Plug-and-play admm for image restoration: Fixed-point convergence and applications,'' {\em IEEE Transactions on Computational Imaging}, vol.~3, no.~1, pp.~84--98, 2016.

\bibitem{romano2017little}
Y.~Romano, M.~Elad, and P.~Milanfar, ``The little engine that could: Regularization by denoising (red),'' {\em SIAM Journal on Imaging Sciences}, vol.~10, no.~4, pp.~1804--1844, 2017.

\bibitem{wang2004image}
Z.~Wang, A.~C. Bovik, H.~R. Sheikh, and E.~P. Simoncelli, ``Image quality assessment: from error visibility to structural similarity,'' {\em IEEE transactions on image processing}, vol.~13, no.~4, pp.~600--612, 2004.

\bibitem{leuschner2021lodopab}
J.~Leuschner, M.~Schmidt, D.~O. Baguer, and P.~Maass, ``Lodopab-ct, a benchmark dataset for low-dose computed tomography reconstruction,'' {\em Scientific Data}, vol.~8, no.~1, p.~109, 2021.

\bibitem{hssayeni2020computed}
M.~Hssayeni, M.~Croock, A.~Salman, H.~Al-khafaji, Z.~Yahya, and B.~Ghoraani, ``Computed tomography images for intracranial hemorrhage detection and segmentation,'' {\em Intracranial hemorrhage segmentation using a deep convolutional model. Data}, vol.~5, no.~1, p.~14, 2020.

\bibitem{karras2017progressive}
T.~Karras, T.~Aila, S.~Laine, and J.~Lehtinen, ``Progressive growing of gans for improved quality, stability, and variation,'' {\em arXiv preprint arXiv:1710.10196}, 2017.

\bibitem{yu2015lsun}
F.~Yu, A.~Seff, Y.~Zhang, S.~Song, T.~Funkhouser, and J.~Xiao, ``Lsun: Construction of a large-scale image dataset using deep learning with humans in the loop,'' {\em arXiv preprint arXiv:1506.03365}, 2015.

\bibitem{nelson2019first}
D.~Nelson, A.~Pillepich, V.~Springel, R.~Pakmor, R.~Weinberger, S.~Genel, P.~Torrey, M.~Vogelsberger, F.~Marinacci, and L.~Hernquist, ``First results from the tng50 simulation: galactic outflows driven by supernovae and black hole feedback,'' {\em Monthly Notices of the Royal Astronomical Society}, vol.~490, no.~3, pp.~3234--3261, 2019.

\bibitem{he2023dynamic}
Z.~He and Z.~Jin, ``Dynamic implicit image function for efficient arbitrary-scale image representation,'' {\em arXiv preprint arXiv:2306.12321}, 2023.

\bibitem{he2024latent}
Z.~He and Z.~Jin, ``Latent modulated function for computational optimal continuous image representation,'' in {\em Proceedings of the IEEE/CVF Conference on Computer Vision and Pattern Recognition}, pp.~26026--26035, 2024.

\bibitem{osato2021}
K.~Osato, J.~Liu, and Z.~Haiman, ``{$\kappa$TNG: effect of baryonic processes on weak lensing with IllustrisTNG simulations},'' {\em Monthly Notices of the Royal Astronomical Society}, vol.~502, pp.~5593--5602, 02 2021.

\bibitem{illustrisTNG1}
A.~{Pillepich}, D.~{Nelson}, L.~{Hernquist}, V.~{Springel}, R.~{Pakmor}, P.~{Torrey}, R.~{Weinberger}, S.~{Genel}, J.~P. {Naiman}, F.~{Marinacci}, and M.~{Vogelsberger}, ``{First results from the IllustrisTNG simulations: the stellar mass content of groups and clusters of galaxies},'' {\em \mnras}, vol.~475, pp.~648--675, Mar. 2018.

\bibitem{illustrisTNG2}
V.~{Springel}, R.~{Pakmor}, A.~{Pillepich}, R.~{Weinberger}, D.~{Nelson}, L.~{Hernquist}, M.~{Vogelsberger}, S.~{Genel}, P.~{Torrey}, F.~{Marinacci}, and J.~{Naiman}, ``{First results from the IllustrisTNG simulations: matter and galaxy clustering},'' {\em \mnras}, vol.~475, pp.~676--698, Mar. 2018.

\bibitem{illustrisTNG3}
J.~P. {Naiman}, A.~{Pillepich}, V.~{Springel}, E.~{Ramirez-Ruiz}, P.~{Torrey}, M.~{Vogelsberger}, R.~{Pakmor}, D.~{Nelson}, F.~{Marinacci}, L.~{Hernquist}, R.~{Weinberger}, and S.~{Genel}, ``{First results from the IllustrisTNG simulations: a tale of two elements - chemical evolution of magnesium and europium},'' {\em \mnras}, vol.~477, pp.~1206--1224, June 2018.

\bibitem{illustrisTNG4}
D.~{Nelson}, A.~{Pillepich}, V.~{Springel}, R.~{Weinberger}, L.~{Hernquist}, R.~{Pakmor}, S.~{Genel}, P.~{Torrey}, M.~{Vogelsberger}, G.~{Kauffmann}, F.~{Marinacci}, and J.~{Naiman}, ``{First results from the IllustrisTNG simulations: the galaxy colour bimodality},'' {\em \mnras}, vol.~475, pp.~624--647, Mar. 2018.

\bibitem{illustrisTNG5}
F.~{Marinacci}, M.~{Vogelsberger}, R.~{Pakmor}, P.~{Torrey}, V.~{Springel}, L.~{Hernquist}, D.~{Nelson}, R.~{Weinberger}, A.~{Pillepich}, J.~{Naiman}, and S.~{Genel}, ``{First results from the IllustrisTNG simulations: radio haloes and magnetic fields},'' {\em \mnras}, vol.~480, pp.~5113--5139, Nov. 2018.

\bibitem{price2020}
M.~A. Price, J.~D. McEwen, L.~Pratley, and T.~D. Kitching, ``{Sparse Bayesian mass-mapping with uncertainties: Full sky observations on the celestial sphere},'' {\em Monthly Notices of the Royal Astronomical Society}, vol.~500, pp.~5436--5452, 11 2020.

\bibitem{kingma2014adam}
D.~P. Kingma and J.~Ba, ``Adam: A method for stochastic optimization,'' {\em arXiv preprint arXiv:1412.6980}, 2014.

\bibitem{dong2018denoising}
W.~Dong, P.~Wang, W.~Yin, G.~Shi, F.~Wu, and X.~Lu, ``Denoising prior driven deep neural network for image restoration,'' {\em IEEE transactions on pattern analysis and machine intelligence}, vol.~41, no.~10, pp.~2305--2318, 2018.

\bibitem{wei2022tfpnp}
K.~Wei, A.~Aviles-Rivero, J.~Liang, Y.~Fu, H.~Huang, and C.-B. Sch{\"o}nlieb, ``Tfpnp: Tuning-free plug-and-play proximal algorithms with applications to inverse imaging problems,'' {\em The Journal of Machine Learning Research}, vol.~23, no.~1, pp.~699--746, 2022.

\bibitem{ahmad2020plug}
R.~Ahmad, C.~A. Bouman, G.~T. Buzzard, S.~Chan, S.~Liu, E.~T. Reehorst, and P.~Schniter, ``Plug-and-play methods for magnetic resonance imaging: Using denoisers for image recovery,'' {\em IEEE signal processing magazine}, vol.~37, no.~1, pp.~105--116, 2020.

\bibitem{wei2020tuning}
K.~Wei, A.~Aviles-Rivero, J.~Liang, Y.~Fu, C.-B. Sch{\"o}nlieb, and H.~Huang, ``Tuning-free plug-and-play proximal algorithm for inverse imaging problems,'' in {\em International Conference on Machine Learning}, pp.~10158--10169, PMLR, 2020.

\bibitem{kamilov2023plug}
U.~S. Kamilov, C.~A. Bouman, G.~T. Buzzard, and B.~Wohlberg, ``Plug-and-play methods for integrating physical and learned models in computational imaging: Theory, algorithms, and applications,'' {\em IEEE Signal Processing Magazine}, vol.~40, no.~1, pp.~85--97, 2023.

\bibitem{kingma2018glow}
D.~P. Kingma and P.~Dhariwal, ``Glow: Generative flow with invertible 1x1 convolutions,'' {\em Advances in neural information processing systems}, vol.~31, 2018.

\bibitem{ho2020denoising}
J.~Ho, A.~Jain, and P.~Abbeel, ``Denoising diffusion probabilistic models,'' {\em Advances in neural information processing systems}, vol.~33, pp.~6840--6851, 2020.

\bibitem{zhu2023denoising}
Y.~Zhu, K.~Zhang, J.~Liang, J.~Cao, B.~Wen, R.~Timofte, and L.~Van~Gool, ``Denoising diffusion models for plug-and-play image restoration,'' in {\em Proceedings of the IEEE/CVF Conference on Computer Vision and Pattern Recognition}, pp.~1219--1229, 2023.

\bibitem{dhariwal2021diffusion}
P.~Dhariwal and A.~Nichol, ``Diffusion models beat gans on image synthesis,'' {\em Advances in neural information processing systems}, vol.~34, pp.~8780--8794, 2021.

\bibitem{boyd2011distributed}
S.~Boyd, N.~Parikh, E.~Chu, B.~Peleato, J.~Eckstein, {\em et~al.}, ``Distributed optimization and statistical learning via the alternating direction method of multipliers,'' {\em Foundations and Trends{\textregistered} in Machine learning}, vol.~3, no.~1, pp.~1--122, 2011.

\bibitem{tan2024provably}
H.~Y. Tan, S.~Mukherjee, J.~Tang, and C.-B. Sch{\"o}nlieb, ``Provably convergent plug-and-play quasi-newton methods,'' {\em SIAM Journal on Imaging Sciences}, vol.~17, no.~2, pp.~785--819, 2024.

\bibitem{gan2024block}
W.~Gan, Y.~Hu, J.~Liu, H.~An, U.~Kamilov, {\em et~al.}, ``Block coordinate plug-and-play methods for blind inverse problems,'' {\em Advances in Neural Information Processing Systems}, vol.~36, 2024.

\bibitem{hauptmann2024convergent}
A.~Hauptmann, S.~Mukherjee, C.-B. Sch{\"o}nlieb, and F.~Sherry, ``Convergent regularization in inverse problems and linear plug-and-play denoisers,'' {\em Foundations of Computational Mathematics}, pp.~1--34, 2024.

\bibitem{pratley2017}
L.~Pratley, J.~D. McEwen, M.~d'Avezac, R.~E. Carrillo, A.~Onose, and Y.~Wiaux, ``{Robust sparse image reconstruction of radio interferometric observations with purify},'' {\em Monthly Notices of the Royal Astronomical Society}, vol.~473, pp.~1038--1058, 09 2017.

\bibitem{liaudat2023}
T.~I. {Liaudat}, M.~{Mars}, M.~A. {Price}, M.~{Pereyra}, M.~M. {Betcke}, and J.~D. {McEwen}, ``{Scalable Bayesian uncertainty quantification with data-driven priors for radio interferometric imaging},'' {\em arXiv e-prints}, p.~arXiv:2312.00125, Nov. 2023.

\bibitem{cohen2016group}
T.~Cohen and M.~Welling, ``Group equivariant convolutional networks,'' in {\em International conference on machine learning}, pp.~2990--2999, PMLR, 2016.

\bibitem{bronstein2021geometric}
M.~M. Bronstein, J.~Bruna, T.~Cohen, and P.~Veli{\v{c}}kovi{\'c}, ``Geometric deep learning: Grids, groups, graphs, geodesics, and gauges,'' {\em arXiv preprint arXiv:2104.13478}, 2021.

\bibitem{cohen2016steerable}
T.~S. Cohen and M.~Welling, ``Steerable cnns,'' {\em arXiv preprint arXiv:1612.08498}, 2016.

\bibitem{veeling2018rotation}
B.~S. Veeling, J.~Linmans, J.~Winkens, T.~Cohen, and M.~Welling, ``Rotation equivariant cnns for digital pathology,'' in {\em Medical Image Computing and Computer Assisted Intervention--MICCAI 2018: 21st International Conference, Granada, Spain, September 16-20, 2018, Proceedings, Part II 11}, pp.~210--218, Springer, 2018.

\bibitem{chaman2021trulyeq}
A.~Chaman and I.~Dokmani{\'c}, ``Truly shift-equivariant convolutional neural networks with adaptive polyphase upsampling,'' in {\em 2021 55th Asilomar Conference on Signals, Systems, and Computers}, pp.~1113--1120, IEEE, 2021.

\bibitem{chaman2021truly}
A.~Chaman and I.~Dokmani{\'c}, ``Truly shift-invariant convolutional neural networks,'' in {\em Proceedings of the IEEE/CVF Conference on Computer Vision and Pattern Recognition}, pp.~3773--3783, 2021.

\bibitem{lowe1999object}
D.~G. Lowe, ``Object recognition from local scale-invariant features,'' in {\em Proceedings of the seventh IEEE international conference on computer vision}, vol.~2, pp.~1150--1157, Ieee, 1999.

\bibitem{puny2021frame}
O.~Puny, M.~Atzmon, H.~Ben-Hamu, I.~Misra, A.~Grover, E.~J. Smith, and Y.~Lipman, ``Frame averaging for invariant and equivariant network design,'' {\em arXiv preprint arXiv:2110.03336}, 2021.

\bibitem{sannai2021equivariant}
A.~Sannai, M.~Kawano, and W.~Kumagai, ``Equivariant and invariant reynolds networks,'' {\em arXiv preprint arXiv:2110.08092}, 2021.

\bibitem{zhao2014rotationally}
Z.~Zhao and A.~Singer, ``Rotationally invariant image representation for viewing direction classification in cryo-em,'' {\em Journal of structural biology}, vol.~186, no.~1, pp.~153--166, 2014.

\end{thebibliography}
\bibliographystyle{ieeetr}

\newpage
\begin{center}
\textbf{\large Supplementary Materials}
\end{center}

\section{Network architecture and training details}
\label{sec: network architecture}
We describe the network architecture and training details of LoFi and CNN baselines used in the experiments in Section \ref{sec: experiments}.

\subsection{Low-dose computed tomography}
For this experiment, we parameterize LoFi with multiMLP architecture depicted in Figure \ref{fig: multiMLP} where we used 9 MLP modules, each with three hidden layers of dimension 370 with ReLu activations and the output layer has 100 neurons. Their outputs are then concatenated and mixed by another MLP with an input layer of size $900 = 9 \times 100$ and 3 hidden layers of dimension 370 with ReLu activations. We consider $K = 81$ neighboring pixels initialized with circular geometry, 9 circles each with 9 points uniformly distributed around the centric coordinate. Regarding the pre-processing filter, we concatenated the real and complex values of the filtered input image along the channel dimension before patch extraction. Regarding the architecture of baselines, we used the standard implementations published by the authors with the following configurations. \\
\textbf{DnCNN:} number of layers=15, number of channels=160 \\
\textbf{IRCNN:} number of layers = 13, number of channels = 256 \\
\textbf{U-Net:} Init-features = 32 \\
\textbf{DRU-Net:} Init-features = 32 \\
\textbf{Restormer:} Transformer blocks = [4, 6, 6, 8], attention heads = [1, 2, 4, 8] and number of channels = [16, 32, 64, 128]. \\
\textbf{SwinIR:} Patch embedding dimension = 120, patch size = 1, window size = 8, attention heads = [6,6,6,6,6,6]. \\

\subsection{Image denoising}
Similar to the experiment for LDCT, we used multiMLP architecture with the same structure. For this particular experiment, we did not use the pre-processing filter to avoid overfitting as our training set is very small. Regarding the architecture of baselines, we used smaller networks for U-Net ahnd DRU-Net for having a fair comparison with other baselines and LoFi in terms of overfitting. \\
\textbf{U-Net:} Init-features = 19 \\
\textbf{DRU-Net:} Init-features = 19 \\
Wit these parameters, all baselines and LoFi have approximately 3M parameres.

\subsection{Dark matter mapping}
\label{app: mass_mapping_app}
For this experiment, we used the same network architecture as the experiment of LDCT for both LoFi and baselines. Regarding dark matter mapping simulations,
we used the $\kappa$TNG convergence maps~\cite{osato2021}, generated from the widely used IllustrisTNG hydrodynamical simulation suite~\cite{illustrisTNG1,illustrisTNG2,illustrisTNG3,illustrisTNG4,illustrisTNG5}, to build the train and test datasets for the models presented in Figure~\ref{fig: results_mass_kapp_128}. The $\kappa$TNG dataset consists of $10\,000$ realizations of $5 \times 5 \text{deg}^{2}$ convergence maps for each of the $40$ different source redshifts with $0 < z_s < 2.6$. For the sake of simplicity, we have selected a single redshift slice, the $20^{\text{th}}$ plane, corresponding to $z_s = 0.858$ for our experiments.

From the simulated convergence fields, $\kappa$, we have constructed the observed shear fields using Equation~\ref{eq:mass_map_forward_op}. We follow~\cite[\S 5.2.1]{price2020} for the noise simulation. The shear noise $\rvn$ for each component of the shear and for each pixel $i$ is simulated as $\rvn_i \sim \mathcal{N}(0, \sigma_i^2)$. The per-pixel standard deviation can be computed as follows
\begin{equation}
    \sigma_i = \frac{\sigma_e}{\sqrt{(\theta^2 / n_{\text{grid}}^2) \times n_{\text{gal}}}},
\end{equation}
where $\sigma_e$ is the standard deviation of the intrinsic ellipticity distribution, $\theta^2$ is the area of the simulated field in arcmin$^2$, $n_{\text{grid}}^2$ is the total number of pixels in the field, and $n_{\text{gal}}$ is the number density of galaxy observations given in number of galaxies per arcmin$^2$. We have used $\sigma_e = 0.37$, which is the typical intrinsic ellipticity standard deviation, $\theta = 300$ arcmin, which corresponds to the $\kappa$TNG simulated area of $5 \times 5 \text{deg}^{2}$, $n_{\text{gal}} = 30$ arcmin$^{-2}$, which corresponds to the projected number density expected in Stage IV surveys like \textit{Euclid}
, and $n_{\text{grid}} = 128$.

\subsection{Training details}
LoFi is implemented in PyTorch on a machine with an Nvidia A100 GPU with 80GB memory. LoFi and baselines were trained for 200 epochs with $\ell_1$ loss using Adam optimizer~\cite{kingma2014adam} with learning rate $10^{-4}$ and object batch size 64. In the case of LoFi, for each mini-batch of random objects, we performed optimization on a random mini-batch of 512 pixels 2 times.

\subsection{Plug-and-Play ADMM}
\label{sec: PnP ADMM}
In this section, we briefly review plug-and-play (PnP) framework for image reconstruction.
PnP algorithms achieved SOTA performance across a variety of image reconstruction tasks including image super-resolution and deblurring~\cite{ dong2018denoising, zhang2021plug}, computed tomography (CT)~\cite{wei2022tfpnp} and magnetic resonance imaging (MRI)~\cite{ahmad2020plug, wei2020tuning}; see also the recent review~\cite{kamilov2023plug}. In PnP, a pre-trained Gaussian denoiser, typically a CNN, is used as a prior. Gaussian denoisers are often cheaper in both than generative priors \cite{kingma2018glow, ho2020denoising} in both training and inference. Training generative models also requires large-scale datasets and substantial computational resources. Recently, the authors of \cite{zhu2023denoising} leveraged diffusion models \cite{ho2020denoising, dhariwal2021diffusion} as a pre-trained denoiser in PnP framework and achieved excellent performance for various image restoration tasks.

Originally, PnP~\cite{venkatakrishnan2013plug} is developed based on the alternating direction method of multipliers (ADMM)~\cite{boyd2011distributed}, where new variables $\rvu$ and $\rvv$ are introduced to decouple the data fidelity and regularization terms in \eqref{eq: forward model} as follows,
\begin{align}
    \min_{\rvf,\rvv} \max_{\rvu} \bigg\{ & \dfrac{1}{2\sigma^2} \| \rvq - \rvA\rvf \|_2^2 + R(\rvv) + \dfrac{1}{2\eta} \| \rvf - \rvv + \rvu \|_2^2 \nonumber \\
    &- \dfrac{1}{2\eta} \| \rvu \|_2^2 \bigg\},
    \label{eq: admm}
\end{align}
where $\eta$ is a penalty parameter that adjusts the convergence rate. We alternate the optimization on $\rvf$, $\rvu$ and $\rvv$ as follows,
\begin{align}
    & \rvf_k = h(\rvv_{k-1}  - \rvu_{k-1}; \alpha), \\
    & \rvv_k = \text{prox}_R(\rvf_k - \rvu_{k-1};\eta), \\
    & \rvu_k = \rvu_{k-1} + (\rvf_k - \rvv_k),
\end{align}
where $\alpha = \dfrac{\sigma^2}{\eta}$ and,
\begin{align}
    & h(\rvz;\alpha) \triangleq (\rvA^* \rvA + \alpha)^{-1} (\rvA^* \rvq + \alpha\rvz), \\
    & \text{prox}_R(\rvz; \eta) \triangleq \argmin_\rvf \dfrac{1}{2\eta} \| \rvf - \rvz \|_2^2 + R(\rvf), \label{eq: proximal_op}
\end{align}
where $\rvA^{*}$ is the adjoint forward operator. 
The proximal operator in~\eqref{eq: proximal_op} can be interpreted as the denoiser of $\rvz$ with image prior $R(\cdot)$ and AWGN variance $\eta$. The key idea of the PnP is to employ a powerful image denoiser, like a pre-trained CNN, for~\eqref{eq: proximal_op}. The convergence analysis of PnP is studied in the literature \cite{tan2024provably, gan2024block, hauptmann2024convergent}.

\subsection{Radio interferometric imaging}
\label{sec: radio}
Radio interferometry is an imaging technique where we acquire samples of the Fourier transform of the image along curves~\cite{pratley2017,liaudat2023}. In radio interferometric imaging, aperture synthesis techniques are used to acquire specific Fourier measurements giving an incomplete coverage of the Fourier plane. The inverse problem consists of recovering the entire intensity image from Fourier measurements. In this case, the forward operator in a simplified form writes
\begin{equation}
    \rvA = \mathbf{M}_{\text{RI}} \mathbf{F}
    \label{eq: radio}
\end{equation}
where $\mathbf{M}_{\text{RI}}$ is a Fourier mask indicating the Fourier coverage of the observations. We have omitted gridding and degridding operations in between other corrections from the forward operator for the sake of simplicity; a more detailed description of the operator can be found in ~\cite[\S 3.2]{pratley2017}.

\begin{figure*}
	\centering
    \includegraphics[width=0.8\linewidth]{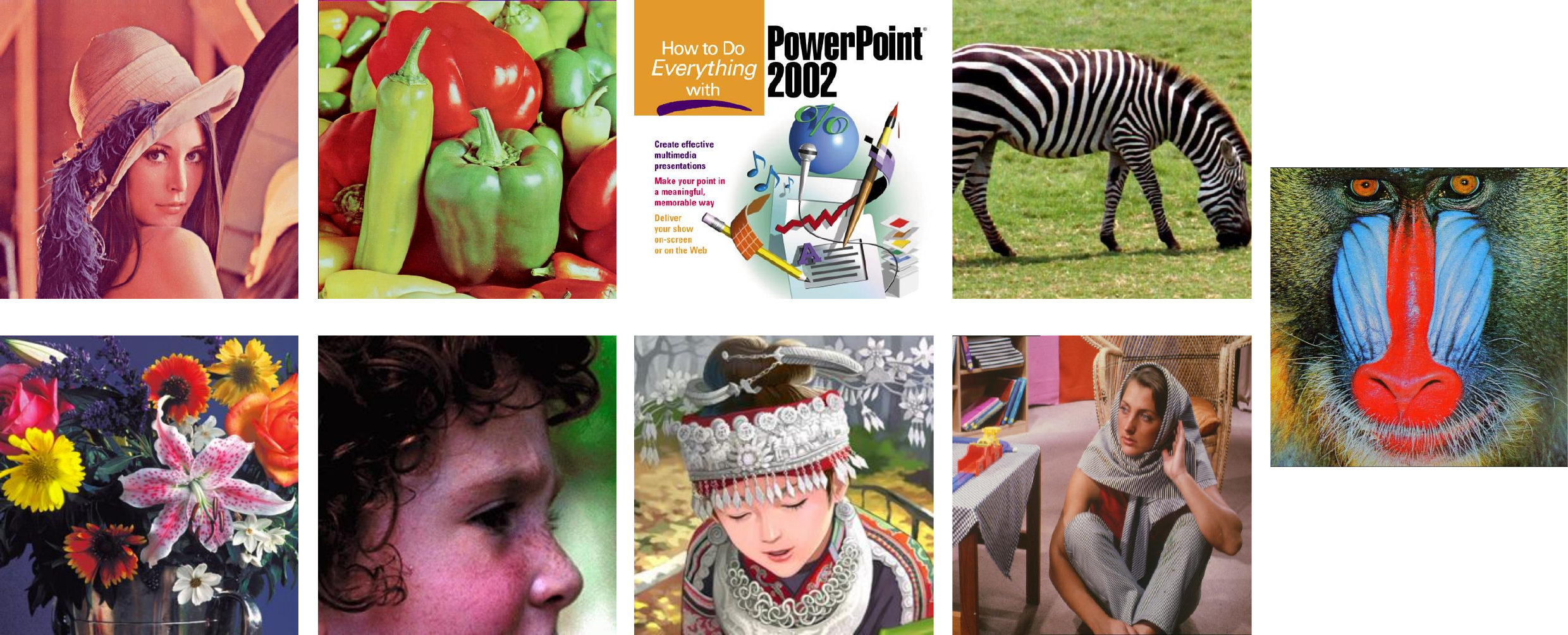}
	\caption{Training dataset used for image denoising experiment at resolution $512 \times 512$.}
	\label{fig: training_set}
\end{figure*}

\section{Futher analysis and additional experiments}
\label{app: additional experiments}
In this section we provide further analysis of LoFi architecture and additional experiments.

\subsection{High-dimensional image denoising}
\label{app: denoising_celeba}
In this section, we train LoFi on 29900 training samples from the CelebA-HQ dataset at resolution $1024 \times 1024$ for image denoising task ($\sigma = 0.2$).

\begin{figure}[t]
    \centering
    \includegraphics[width = 0.45\textwidth]{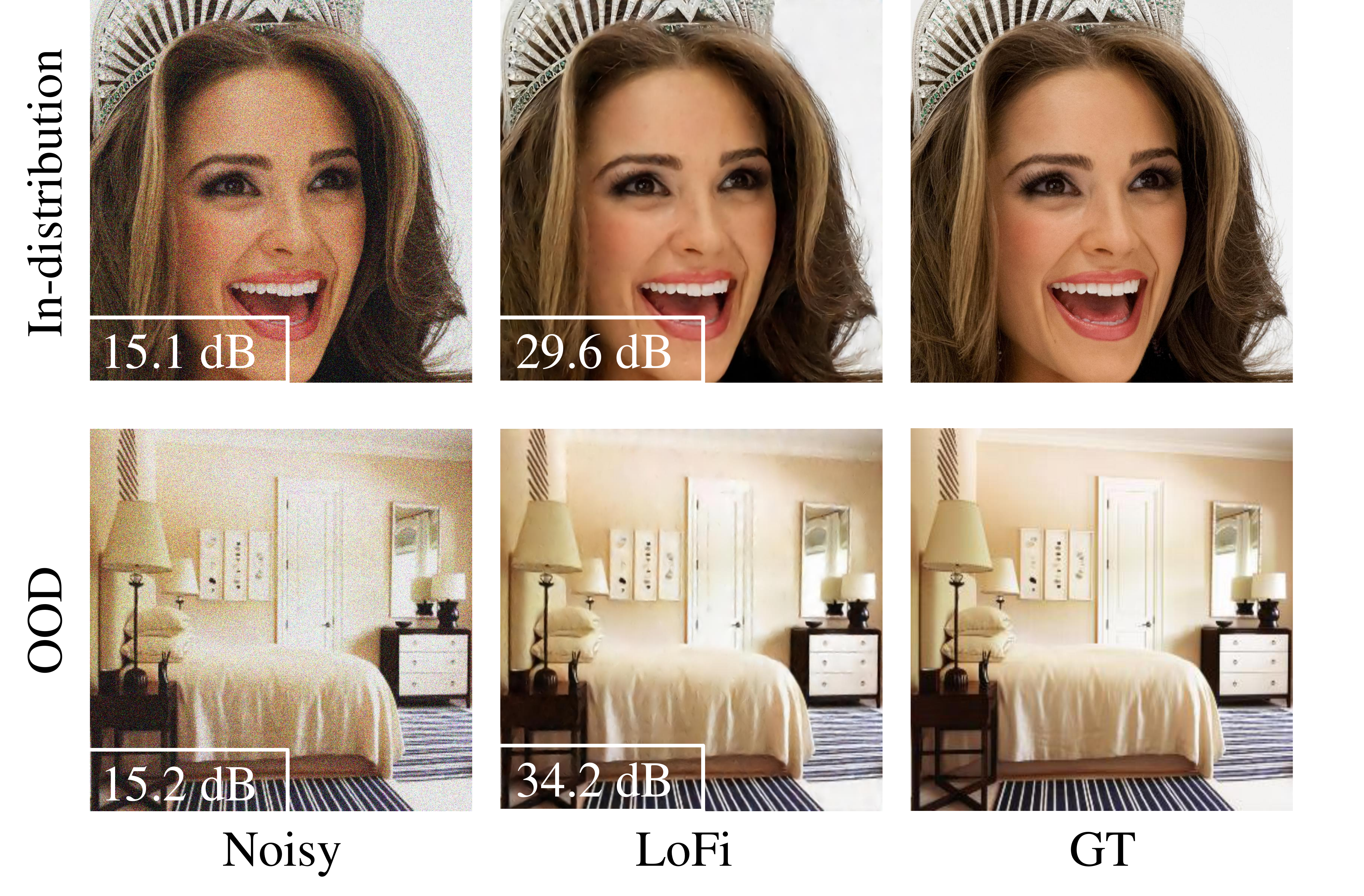}
    \\ [-5pt]
    \caption{LoFi reconstructions on in-distribution and OOD data for image denoising ($\sigma = 0.2$) at resolution $1024 \times 1024$.}
    \label{fig: results_denoising_celeba_1024}
\end{figure}

Figure~\ref{fig: results_denoising_celeba_1024} illustrates the reconstructed images for both in-distribution and OOD samples. This experiment shows that LoFi can achieve high-quality reconstructions for large images despite the simplicity of its architecture.

\subsection{Equivariance to shifts and rotations}
\label{sec: rot_equi}
In many imaging and pattern recognition tasks, it is desirable to work with functions that are invariant or equivariant to shifts and rotations. Building such invariances into CNNs, typically via group representation theory~\cite{cohen2016group, bronstein2021geometric} improves generalization both in and out of distribution~\cite{cohen2016steerable, veeling2018rotation}.  

It is easy to see that LoFi is shift equivariant by design. Indeed, the patch at position $\rvx + \Delta \rvx$ in an image shifted by $\Delta \rvx$ is exactly the same as the patch at position $\rvx$ in a non-shifted image; the $\text{NN}_\theta$ (MultiMLP) thus receives the same input.
This stands in contrast to most mainstream multi-scale CNNs which are surprisingly sensitive even to 1-pixel shifts, although equivariance can be restored by a careful design of pooling and downsampling layers \cite{chaman2021trulyeq, chaman2021truly}.

Rotation equivariance is precluded for square patches, except for angles that are multiples of $\frac{\pi}{2}$. With circular patches shown in Figure~\ref{fig: multiMLP}, when the image is rotated the $\text{NN}_\theta$ receives the same input only in a different order up to interpolation error. Rotation equivariance could be implemented by applying $\text{NN}_\theta$ to rotation invariant features~\cite{lowe1999object, puny2021frame, sannai2021equivariant, zhao2014rotationally} extracted from the circular patch or by averaging the output over rotations. Even without this, however, LoFi is approximately rotation equivariant because natural images are approximately \textit{locally} rotation invariant and the patch geometry is initialized as circular.

\begin{figure}[t]
    \centering
    \includegraphics[width = 0.5\textwidth]{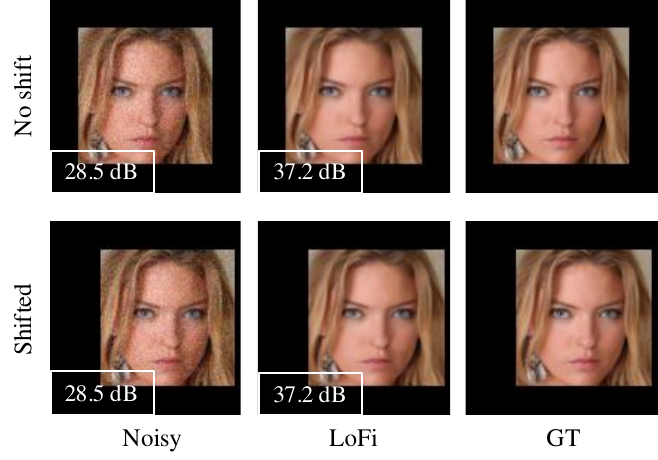}
    \\ [-5pt]
    \caption{Shift equivariance analysis for LoFi architecture; LoFi exhibits no degradation in reconstruction quality when the input noisy image is shifted for 15 pixels.}
    \label{fig: shift_equi}
\end{figure}

Now, we assess LoFi performance on the image denoising task when the input image is shifted or rotated. As shown in Figure~\ref{fig: shift_equi}, we horizontally shifted the input noisy image for 15 pixels. As expected, LoFi's reconstruction is shifted for 15 pixels and exhibits no degradation in the reconstruction quality. This experiment confirms that LoFi is a truly shift equivariant network.

\begin{figure}[t]
    \centering
    \includegraphics[width = 0.5\textwidth]{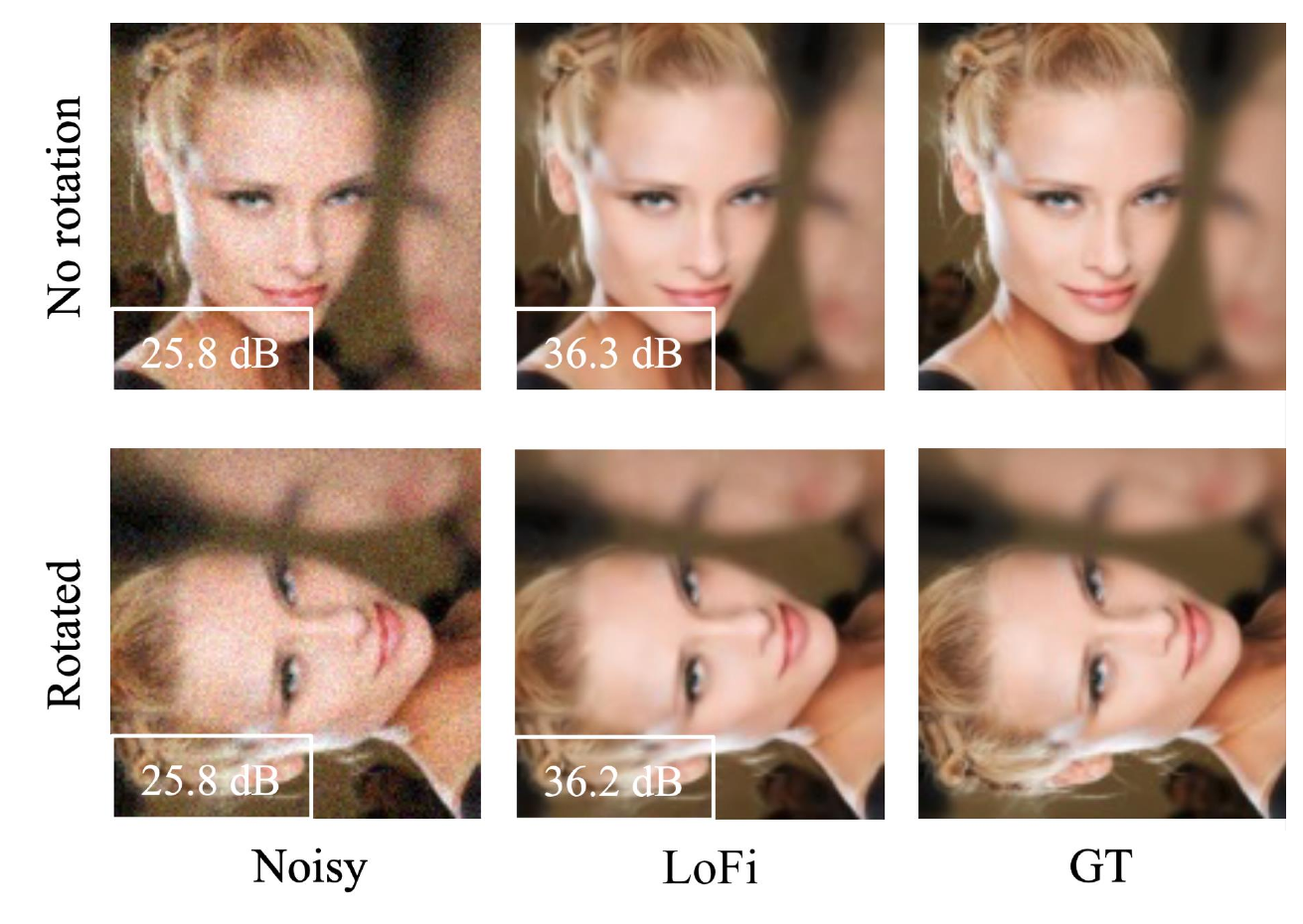}
    \\ [-5pt]
    \caption{Rotation equivariance analysis for LoFi architecture; LoFi exhibits minimal degradation in reconstruction quality when the input noisy image undergoes a 90-degree rotation.}
    \label{fig: rot_equi}
\end{figure}

Figure~\ref{fig: rot_equi} demonstrates the LoFi performance when the input noisy image undergoes a 90-degree rotation. As expected, we observe minimal degradation in reconstruction quality which shows LoFi's strong robustness under rotation. Note that, we did not use any data augmentation to make the model robust under translation and rotation during training.  

\end{document}